\documentclass[10pt,journal,compsoc]{IEEEtran}
\usepackage{graphicx}
\usepackage{amsmath}
\usepackage{amssymb}
\usepackage{framed,multirow}
\usepackage{array}
\usepackage{ragged2e}
\usepackage{setspace}
\usepackage{amsmath}
\usepackage{color}
\usepackage[pagebackref=true,colorlinks,bookmarks=false,citecolor=blue,linkcolor=blue,urlcolor=blue]{hyperref}
\usepackage[table]{xcolor}    % loads also »colort
\definecolor{lightgray}{gray}{0.9}
\ifCLASSOPTIONcompsoc
\usepackage[nocompress]{cite}
\else
\usepackage{cite}
\fi

\begin{document}
	
	\title{Learning Knowledge-based Prompts for\\ Robust 3D Mask Presentation Attack Detection}
	
	\author{Fangling~Jiang, Qi~Li,  Bing Liu,  Weining Wang, Caifeng Shan, Zhenan~Sun, Ming-Hsuan Yang% <-this % stops a space
		\IEEEcompsocitemizethanks{
        	\IEEEcompsocthanksitem This work was supported in part by the National Key Research and Development Program of China (Grant No. 2022YFC3310400), the Natural Science Foundation of China (Grant Nos. U23B2054, 62102419, 62276263 and 62406133), the Ministry of Education Research Foundation (Grant No. 23YJA630058), the Natural Science Foundation of Hunan Province (Grant Nos.2024JJ6389 and 2024JJ7437), the Hengyang Science and Technology Plan Project (Grant No.202330046190), and the Interdisciplinary Research Program in Medicine and Engineering, The First Affiliated Hospital of University of South China (Grant No. IRP-M$\&$E-2025-12). (Corresponding author: Qi~Li.)
			\IEEEcompsocthanksitem Fangling Jiang and Qi Li contributed equally to this study.  
			\IEEEcompsocthanksitem Fangling Jiang and Bing Liu are with the School of Computer Science, University of South China, Hengyang, 421001, China (email: \{jfl, bingliu\}@usc.edu.cn). 
			\IEEEcompsocthanksitem Qi Li, Weining Wang and Zhenan Sun are with the New Laboratory of Pattern Recognition, MAIS, CASIA, Beijing, 100190, China (email: \{qli, weining.wang, znsun\}@nlpr.ia.ac.cn). 
			\IEEEcompsocthanksitem Caifeng Shan is with the School of Intelligence Science and Technology, Nanjing University, Suzhou, 215163, China (email:caifeng.shan@gmail.com). 
			\IEEEcompsocthanksitem Ming-Hsuan Yang is with the Department of Computer Science and Engineering, University of California, Merced, CA, 95340, USA (email:mhyang@ucmerced.edu), and Department of Computer Science and Engineering, Yonsei University, Korea. 
		}% <-this % stops an unwanted space
	}
	
	\IEEEtitleabstractindextext{%
		\begin{abstract}
			\justifying
			3D mask presentation attack detection is crucial for protecting face recognition systems against the rising threat of 3D mask attacks. While most existing methods utilize multimodal features or remote photoplethysmography (rPPG) signals to distinguish between real faces and 3D masks, they face significant challenges, such as the high costs associated with multimodal sensors and limited generalization ability. 
			Detection-related text descriptions offer concise, universal information and are cost-effective to obtain. However, the potential of vision-language multimodal features for 3D mask presentation attack detection remains unexplored. 
			In this paper, we propose a novel knowledge-based prompt learning framework to explore the strong generalization capability of vision-language models for 3D mask presentation attack detection. Specifically, our approach incorporates entities and triples from knowledge graphs into the prompt learning process, generating fine-grained, task-specific explicit prompts that effectively harness the knowledge embedded in pre-trained vision-language models. Furthermore, considering different input images may emphasize distinct knowledge graph elements, we introduce a visual-specific knowledge filter based on an attention mechanism to refine relevant elements according to the visual context. Additionally, we leverage causal graph theory insights into the prompt learning process to further enhance the generalization ability of our method. During training, a spurious correlation elimination paradigm is employed, which removes category-irrelevant local image patches using guidance from knowledge-based text features, fostering the learning of generalized causal prompts that align with category-relevant local patches. Experimental results demonstrate that the proposed method achieves state-of-the-art intra- and cross-scenario detection performance on benchmark datasets. 
		\end{abstract}        
		}
	
	\maketitle
	
	\IEEEdisplaynontitleabstractindextext
	
	\IEEEpeerreviewmaketitle
	\IEEEraisesectionheading{\section{Introduction}\label{sec:introduction}}
	
	\IEEEPARstart{F}{ace} recognition systems, known for their high accuracy and convenience, have been deeply integrated into everyday applications, including residential access control, attendance systems, and criminal tracking. However, face presentation attacks pose a significant challenge to the reliability of these systems~\cite{ramachandra2019custom}. Imposers commonly exploit printed photos, replayed videos, or 3D masks to impersonate real faces and deceive face recognition systems. To counter these threats, the task of face presentation attack detection (PAD), which aims at distinguishing fake faces from real ones, has drawn considerable attention from both academia and industry~\cite{yu2021deep,dharmawan2024towards}.%de2013can,
	
	Most face presentation attack detection methods predominantly focus on identifying 2D spoof faces, such as printed photos and replayed videos~\cite{yu2021deep,dharmawan2024towards}. Over the past decade, these methods have utilized either handcrafted features or deep learning-based representations, achieving commendable detection performance~\cite{mmboulkenafet2015face,hu2024rethinking,yang2024generalized}. In contrast, 3D masks pose a greater challenge due to their high similarity to real faces in color, texture, and three-dimensional structure. Advances in manufacturing technologies, including 3D printing, have significantly reduced the production cost of 3D masks while increasing their accessibility. As a result, the threat of 3D mask attacks has grown substantially in recent years. However, numerous discriminative features effective for 2D spoof detection, such as image blurring, blinking patterns, and depth map analysis, have been proved inadequate for 3D mask detection. To address these challenges, researchers have initiated specialized studies to tackle the difficulties of 3D mask presentation attack detection~\cite{jia2020survey,liu2023recent}.
	
	Motivated by 2D spoof face detection, numerous methods have employed global and local facial texture features~\cite{grinchuk20213d,liu2022contrastive} or motion cues derived from optical flow maps~\cite{shao2017deep,cao2024flow} to differentiate 3D masks from real faces. Additionally, since real faces exhibit heartbeat signals while 3D masks do not, rPPG signals, which show blood volume pulse, have become a widely used technique for 3D mask presentation attack detection in recent studies~\cite{li2016generalized,yao2023mask,yu2021transrppg}.
	Although these methods have shown promising detection performance in specific scenarios, they are still sensitive to changes in lighting, recording devices, and subtle facial movements~\cite{jia2020survey}. 
	Considering the different materials of 3D masks and real faces,  images across multiple spectral bands, from visible light to long-infrared wavelength have been collected~\cite{kim2009masked,sun20203d,kotwal2019multispectral} for this problem.  
	The aim is to exploit the reflective differences in these multimodal images to distinguish between real faces and 3D masks. Although multimodal features-based methods are highly resistant to environmental variations, they require specialized equipment, such as near-infrared or thermal infrared cameras, to capture non-visible spectrum images. The high cost of such devices often limits their practicality in real-world applications.
	
\begin{figure}[tp]
		\centering
		\includegraphics[width=0.5\textwidth]{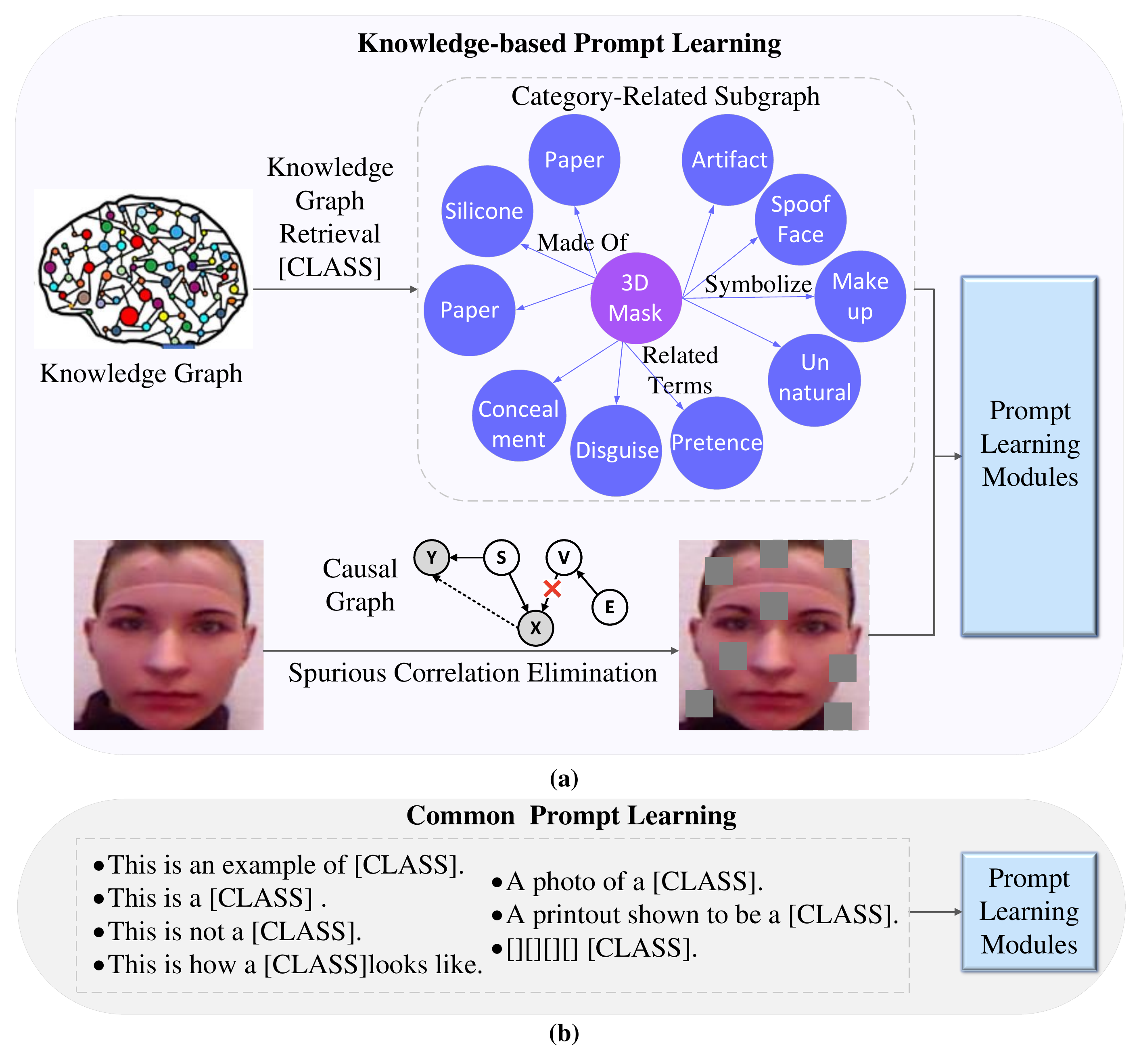}
		\caption{(a) The proposed knowledge-based prompt learning paradigm explicitly integrates expert knowledge in knowledge graphs and causal representation learning for prompt learning. 
			(b)	Common prompt learning only includes category names for business-related contexts.
		}
		\label{fig1}
	\end{figure}

	Pre-trained vision-language models have recently been effectively applied to downstream visual tasks, demonstrating exceptional generalization performance~\cite{zhang2024vision}. Despite this success, their application to 3D mask presentation attack detection remains unexplored. Given that category-related text encapsulates concise and universal information, pre-trained vision-language models can enhance 3D mask presentation attack detection by integrating discriminative and generalized features from multiple modalities at a reduced cost.
	In this paper, we propose a novel knowledge-based prompt learning framework, which explores well-generalized pre-trained vision-language models for 3D mask presentation attack detection through optimizations in both prompt scheme and learning strategies, as illustrated in \figurename~\ref{fig1}.
	
	In this work, we integrate both entities and triples from knowledge graphs into the prompt generation process.  In the context of 3D mask presentation attack detection, task-specific categories like ``real faces'' carry semantically rich information. However, generic category names or common text prompts (e.g., ``a photo of [CLASS]'') fail to convey this nuanced knowledge~\cite{Zhou_2022,zhou2022conditional,srivatsan2023flip,fang2024vl}.  Knowledge graphs, in contrast, encapsulate extensive expert knowledge, providing rich task-specific contextual information. To leverage this potential, we propose explicitly incorporating both knowledge graph entities and fine-grained discriminative descriptions derived from knowledge graph triples into prompt learning.
	Specifically, we retrieve subgraphs related to category names from common knowledge graphs such as ConceptNet~\cite{speer2017conceptnet} and augment the category names with experiential knowledge by extracting relevant entities and triples. Given that different subgraph elements may vary in relevance across input images, we design a visual-specific knowledge filter based on an attention mechanism to dynamically select pertinent knowledge according to the visual representation. This approach generates context-aware knowledge graph prompts tailored to the task at hand.
	
	Furthermore, we integrate expert insights from causal representation learning into the prompt-learning strategy to improve the generalization capability of our algorithm in cross-scenario 3D mask presentation attack detection. 
	Causal representation learning highlights that poor cross-scenario generalizability in numerous detection models is often due to spurious correlations between categories and scenario-specific, category-irrelevant factors (e.g., backgrounds). 
	To address this issue, we introduce a spurious correlation elimination learning paradigm, which leverages knowledge-based text features to explicitly identify and filter out category-irrelevant local image patches. This approach enables the learning of generalized causal prompts that align more effectively with category-relevant local patches, significantly enhancing model robustness across diverse scenarios.
	
	The main contributions of this paper are:
	\begin{enumerate}
		\item  We propose a novel knowledge-based prompt learning framework, which explores well-generalized pre-trained vision-language models for 3D mask presentation attack detection through optimizations in both prompt schemes and learning strategies.
		\item  We incorporate both knowledge graph entities and fine-grained discriminative descriptions derived from knowledge graph triples into prompt learning to leverage the expert domain knowledge from knowledge graphs.
		\item We propose a spurious correlation elimination prompt learning paradigm, which leverages causal prompt learning to explicitly identify and remove category-irrelevant local image patches to improve the generalization capability.
		\item Extensive experimental results demonstrate that our method achieves state-of-the-art intra- and cross-scenario 3D mask presentation attack detection performance on the benchmark datasets, including high-fidelity masks with color, texture, and structural patterns never encountered in the training set.
	\end{enumerate}

	\section{Related Work}
        In this section, we review related work from three perspectives: 3D mask presentation attack detection, vision-language model-based attack detection, and evaluation metrics commonly used for face presentation attack detection.
	\subsection{3D Mask Presentation Attack Detection}
        Based on the modality of the input image, we categorize 3D mask presentation attack detection methods into two main groups: single-modality and multimodality-based approaches.
	\subsubsection{Single-Modality-based Detection Methods}
	
	\textbf{Texture-based methods.}
	3D masks simulate more three-dimensional structures than printed photos and replayed videos. However, due to the manufacturing technology, common 3D masks still have some color distribution and texture defects.
	Multi-scale LBP features and the SVM classifier are combined to explore the texture disparity for real face and 3D mask classification~\cite{kose2013countermeasure,liu20163d}.
	In contrast to handcrafted features, the vanilla and central difference convolution streams are joined to learn comprehensive deep features for 3D mask presentation attack detection in~\cite{chen2021dual}. 
	Compared to plastic and resin masks, silicone masks have more realistic textures, which require more effective methods to extract texture features.
	Manjani et al.~\cite{manjani2017detecting} use a multilevel deep dictionary to detect high-fidelity silicone masks.
	In~\cite{agarwal2019chif}, Agarwal et al. extract histogram image features from images convolved with non-linearly learned filters to capture the texture better.
	Recently, Grinchuk et al.~\cite{grinchuk20213d} split face images into multiple face parts, and then learn deep features from local patches, which is shown to be conducive to mining microtexture differences.
	Contrastive context-aware learning is performed to exploit rich contextual cues for 3D mask presentation attack detection in~\cite{liu2022contrastive}.
	
	\noindent \textbf{Motion-based methods.}
	Due to the material's lack of softness, most current masks still have difficulty simulating facial motions such as facial expressions and mouth movements.
	Optical flow features are usually used to capture facial motion disparity.
	Siddiqui et al.~\cite{siddiqui2016face} fuse features extracted by the histogram of oriented optical flows and multi-LBP to discriminate the real faces and 3d masks. 
	Shao et al.~\cite{shao2017deep} explore subtle facial motions via estimating optical flow from convolutional feature channels and learning dynamic texture features with a channel-discriminability constraint.
	A flow attention network is designed to mine inter- and intra-frame optical flow features to acquire fine-grained motion details for 3D mask detection in~\cite{cao2024flow}, improving the generalization ability on mask materials, camera sensors, and environmental conditions.
	
	\noindent \textbf{rPPG-based methods.} 
	Remote photoplethysmography (rPPG) is a technique that uses sensors such as visible light cameras to capture periodic changes in skin color caused by the cardiac cycle. The rPPG technology can extract blood volume pulse signals to measure heart rate.
	Real faces have pulse signals, but 3D masks do not. This significant difference can be used to distinguish between real faces and 3D masks~\cite{jia2020survey}.
	Li at al.~\cite{li2016generalized} compute rPPG signals from local parts of the lower half face, and extract six-dimensional handcraft features for 3D mask presentation attack detection.
	Multi-scale long-term statistical spectral features are extracted from rPPG signals for 3D mask and real face classification in~\cite{lin2019face}.
	
	External lighting conditions, recording equipment, facial micro-movements, and face alignment would introduce noise to the rPPG signals.
	To reduce the negative impact of noise, Liu et al.~\cite{liu2018remote} introduce global noise into spectrum template learning and extract the correspondence between the learned spectrum template and the local rPPG signals as discriminative features. 
	Further, Liu et al.~\cite{liu2020temporal} extract the temporal similarity of local rPPG signals between real faces and 3D masks regarding the amplitude, gradient, and phase. This improves the robustness of the detection model to external multi-factor variations on one hand and reduces the need of the input video length on the other.
	Unlike previous methods that manually design feature descriptors to extract discriminative features from rPPG signals, Yu et al.~\cite{yu2021transrppg} take a data-driven perspective and propose a transformer model to learn features from rPPG-based multi-scale spatial-temporal maps automatically.
	Moreover, Yao et al.~\cite{yao2023mask} strengthen the robustness of the detection model to facial motion in terms of face alignment, rPPG signal extraction, and feature learning network. They fuse the rPPG signals on multiple color channels and design a lightweight EfficientNet to learn both spatial and temporal features for real face and 3D mask classification.
	
	\subsubsection{Multimodality-based Detection Methods}
	Common materials used to make 3D masks are paper, plastic, resin, latex, and silicone. These materials are significantly different from those of human faces. Different materials lead to an optical characteristic disparity between real faces and 3D masks, especially in spectra other than the visible spectrum~\cite{jia2020survey}. 
	Several approaches fuse different information embedded in multispectral images for 3D mask presentation attack detection.
	Kim et al.~\cite{kim2009masked} first compare albedo between real faces and 3D mask materials using radiance measurements of the forehead region under the 850 and 685 nm dual wavelengths. Subsequently, wavelengths from visible light to long-wave infrared have been used for 3D mask presentation attack detection, such as polarization medium wave infrared images~\cite{sun20203d}, the fusion of visible and near-infrared features~\cite{liu2018detecting}, and the fusion of visible, near-infrared, and thermal infrared features~\cite{bhattacharjee2017you,george2019biometric,agarwal2017face,kotwal2019multispectral}.
	Multi-channel convolutional neural networks are also used to fuse the representations of multi-spectral images~\cite{george2019biometric,kotwal2019multispectral}.
	
	In addition to multispectral images, depth maps, which contain rich 3D structural information disparity, are often used for 3D mask presentation attack detection~\cite{raghavendra2014novel,george2019biometric,erdogmus2014spoofing,kose2013shape}.
	For instance, ~\cite{raghavendra2014novel} extracts discriminative features from visible light and depth maps regarding both local and global perspectives. The local level focuses on sharp and discontinuous features in regions rich in imitation cues, such as eyes and nose. The global level, on the other hand, uses binarized statistical image features to extract global microtexture differences.
	Besides, Erdogmus et al.~\cite{erdogmus2014spoofing,kose2013shape} extract LBP features to capture the texture and 3D structural disparity for 3d mask detection.
	
	Multimodality-based methods require professional equipment such as near-infrared, thermal infrared, or depth cameras to capture multimodal images. These devices will increase the cost of 3D mask presentation attack detection. This will also limit the application of such methods to a certain extent.
	The proposed method learns multimodal features from text and visible light images. Compared to other modalities requiring additional equipment, the text modality is cost-effective and does not require extra devices. Moreover, the fine-grained knowledge graph prompts encapsulate rich and universal knowledge, enhancing the discriminative power and generalization ability for 3D mask presentation attack detection.
	
	\subsection{Vision-Language Models for Attack Detection}
	
	Visual-language models are multimodal models that learn from both images and text. Large visual-language models have good zero-shot generalization capability~\cite{park2023visual}.
	Since large-scale data collection is difficult for downstream visual tasks, exploring the pre-trained knowledge of the foundation model for these tasks is a more efficient approach than training from scratch.
	Prompt learning, which focuses on developing optimal context descriptions to bridge the gap between models and downstream tasks, is a commonly used method~\cite{zhang2024vision}. 
	
	In light of recent advances in natural language processing~\cite{liu2021pretrain}, numerous methods focus on text prompt learning. For example, CoOp~\cite{Zhou_2022} introduces learnable word vectors to capture context information for various classes. CoCoOp~\cite{zhou2022conditional} builds upon CoOp, generating image-specific text prompts through a meta-learning approach to reduce overfitting on known classes. Unlike text prompt learning, VPL~\cite{bahng2022exploring} involves learning an image perturbation added to input images as visual prompts. Additionally, unified prompt tuning~\cite{zang2022unified} and multi-modal prompt learning~\cite{khattak2023maple} integrate text and visual prompt learning simultaneously, combining both benefits. Recently, prompt learning has been applied to advanced vision task face presentation attack detection~\cite{srivatsan2023flip,yu2023visual,fang2024vl,liu2024cfpl,guo2024style,mu2023teg}.
	
	Srivatsan et al.~\cite{srivatsan2023flip} design a collection of fixed category prompts for real and fake faces and fine-tune all parameters of the CLIP model to align image and text features for face presentation attack detection.
	Yu et al.~\cite{yu2023visual} learn the modal-relevant visual prompts to explore the ability of the frozen foundation model for the problems of partial modal absence in multimodality-based face presentation attack detection.
	Meanwhile, textually prompt learning has been introduced for domain generalization-based face presentation attack detection in~\cite{fang2024vl,liu2024cfpl,mu2023teg,guo2024style}.
	Fang et al.~\cite{fang2024vl} provide fine-grained textual descriptions of the face region to reduce the model's focus on irrelevant facial information and develop sample-level image-text alignment strategies to learn domain-invariant features.
	Different semantic prompts conditioned on content and style features are learned by two lightweight transformers in~\cite{liu2024cfpl}.
    Similarly, the learnable prompt tokens are required to carry diverse visual styles from implicit synthesis of mixed novel styles in~\cite{guo2024style}, enhancing the generalizability of the learned prompts to domain-specific stylistic variations.
	Additionally, Mu et al.~\cite{mu2023teg} construct matching and non-matching textual prompts and use these cross-domain generic textual prompts to explicitly constrain the model focusing on domain-invariant features rather than domain-specific details such as lighting conditions, and recording devices.

	In contrast to existing prompt learning-based face presentation attack detection approaches that primarily focus on detecting 2D spoof faces, our method targets the more challenging task of high-fidelity 3D mask detection. These masks exhibit high similarity to real faces in terms of color, texture, and structure, making them significantly more difficult to distinguish.
    To address this challenge, we inject explicit and interpretable knowledge graphs into prompt learning, while constructing a spurious correlation elimination learning paradigm to jointly adapt general knowledge of foundation models for the detection task.
	
        \subsection{Evaluation Metrics for Face Presentation Attack Detection}
	The performance evaluation of face presentation attack detection algorithms requires a comprehensive assessment of error rates associated with both real faces and presentation attacks. The evaluation metrics commonly used in this field can be broadly categorized into two principal groups:
	
	The first category comprises traditional biometric classification metrics~\cite{chingovska2012effectiveness}, including False Acceptance Rate (FAR), False Rejection Rate (FRR), Equal Error Rate (EER), Half Total Error Rate (HTER), and Area Under the Receiver Operating Characteristic Curve (AUC).
	FAR refers to the proportion of presentation attacks incorrectly accepted as real, while FRR denotes the proportion of real faces incorrectly rejected as presentation attacks. By varying the decision threshold, pairs of FAR and FRR values are obtained, forming the Receiver Operating Characteristic (ROC) curve.
	EER is calculated on the development set and corresponds to the point on the ROC curve where FAR equals FRR. It reflects the threshold at which the false acceptance and false rejection rates are balanced.
	HTER is computed on the test set using the decision threshold determined on the development set (i.e., at EER), and is defined as the average of FAR and FRR on the test set.
	AUC quantifies the area under the ROC curve and serves as a comprehensive measure of overall classification performance.
	
	The second category includes standardized metrics proposed by ISO/IEC 30107-1~\cite{evabz}, namely: Bonafide Presentation Classification Error Rate (BPCER), Attack Presentation Classification Error Rate (APCER), and Average Classification Error Rate (ACER).
	BPCER captures the error rate associated with misclassifying real presentations. In contrast to FAR, which aggregates all attack types into a single category, APCER evaluates each attack type individually and reports the highest error rate among them, thus reflecting the vulnerability of models to the most challenging attack type.
	ACER is defined as the average of APCER and BPCER, providing a balanced evaluation that considers both false acceptances and false rejections equally.

	\begin{figure*}[!t]
		\centering
		\includegraphics[width=1\textwidth]{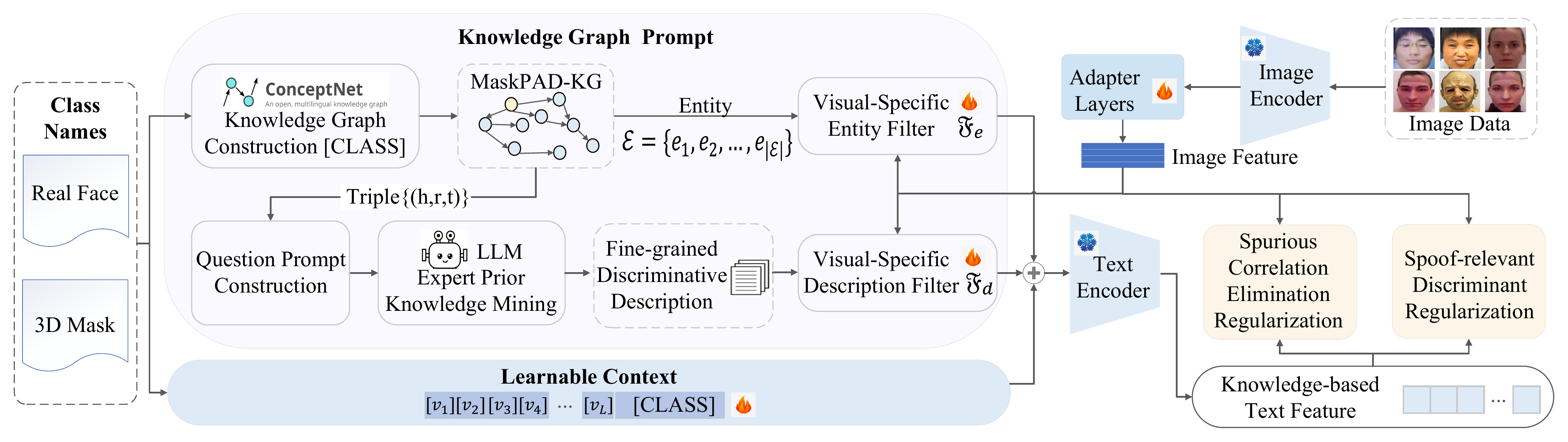}
		\caption{The proposed framework comprises two key components: a knowledge graph prompt module and a spurious correlation elimination module. The knowledge graph prompt module integrates entities and triples from the knowledge graph, along with learnable contexts, as fine-grained prompts to effectively adapt pre-trained vision-language models for 3D mask presentation attack detection. Meanwhile, the spurious correlation elimination module employs causal prompt learning to explicitly identify and remove category-irrelevant local image patches, thereby enhancing the model's generalization capability.		
		}
		\label{fig2}
	\end{figure*}

	\section{Proposed Method}
	Denote the available training domain as  
	$\mathcal{D}=\left\{\left(x_{i}, y_{i}\right)\right\}_{i=1}^{N}$, where
	$x_{i}$ represents the $i$-th data sample, and $y_{i}$ is its corresponding label, belonging to the label set $\mathcal{C} \in \left\{ {1,\ldots,K} \right\}$, and $K$ denotes the total number of classes.
	The label set $\mathcal{C}$ includes two classes: real faces and 3D masks.
	Our objective is to learn a 3D mask presentation attack detection model trained on $\mathcal{D}$, with the capability to generalize effectively to an unseen testing domain $\mathcal{D}^{u}=\left\{x_{i}^{u}\right\}_{i=1}^{N^u}$.
	The testing domain shares the same label set with the training domain but may exhibit a different data distribution due to variations in factors such as recording scenes, devices, lighting conditions, and the appearance or material of 3D masks. 
	To achieve this objective, we propose a knowledge-based prompt learning framework, as illustrated in~\figurename~\ref{fig2}.

	\subsection{Knowledge Graph Prompt Scheme}
	\label{32}
	Common knowledge graphs, such as ConceptNet and Wikidata~\cite{vrandevcic2014wikidata}, encapsulate a wealth of domain expert knowledge. This knowledge is typically a comprehensive and well-organized summary of experts from a large collection of samples. For instance, mask materials may include paper, plastic, silicone, and resin. Such extensive and structured knowledge undoubtedly aids in learning generalized features for 3D mask presentation attack detection. We design a knowledge graph prompt tailored for the vision-language model to leverage this.
	
	 \noindent \textbf{MaskPAD-KG construction.} 
	We begin by using category names in the label set $\mathcal{C}$ to query common public knowledge graphs such as ConceptNet and Wikidata, retrieving category-relevant entities and relationships. To ensure the task relevance and accuracy of the resulting subgraph, we manually inspect the retrieved results and remove entities or relationships unrelated to the 3D mask presentation attack detection task (e.g., the entity \textit{batman} associated with \textit{3D mask}).
        In this work, prior knowledge is organized in three aspects:
    category-related terms, which frequently co-occur with the category name in contextual descriptions (e.g., \textit{pretense}, \textit{concealment});
    symbolic meanings, which represent abstract meanings or synonyms assigned to the category by humans in the context of 3D mask presentation attack detection (e.g., \textit{unreal}, \textit{unnatural});
    and inherent characteristics, which describe the essential properties of the category (e.g., \textit{uniform texture}, \textit{silicone material}).
    These three dimensions allow the knowledge graph to extend beyond surface-level correlations and capture deeper semantic attributes of each category, thereby enriching the category names with more meaningful context.
    
	While most general-purpose knowledge graphs focus on broad commonsense knowledge, they may lack some task-specific details critical for 3D mask presentation attack detection. To address this, we augment the retrieved subgraph by incorporating additional entities based on domain-specific cues commonly observed in 3D mask presentation attack detection. The augmented entities are similarly organized into the aforementioned three semantic dimensions. For example, for the 3D mask category, we add symbolic term \textit{spoof face}, and inherent characteristics including \textit{edge}, \textit{seam}, \textit{subtle contour}, and \textit{rigid shape}.
	Overall, the effort required to construct the knowledge graph is minimal, as it is only performed for two categories: real face and 3D mask.
	
\begin{figure*}[!t]
		\centering
            %\captionsetup{justification=centering,labelsep=period}
		\includegraphics[width=0.9\textwidth]{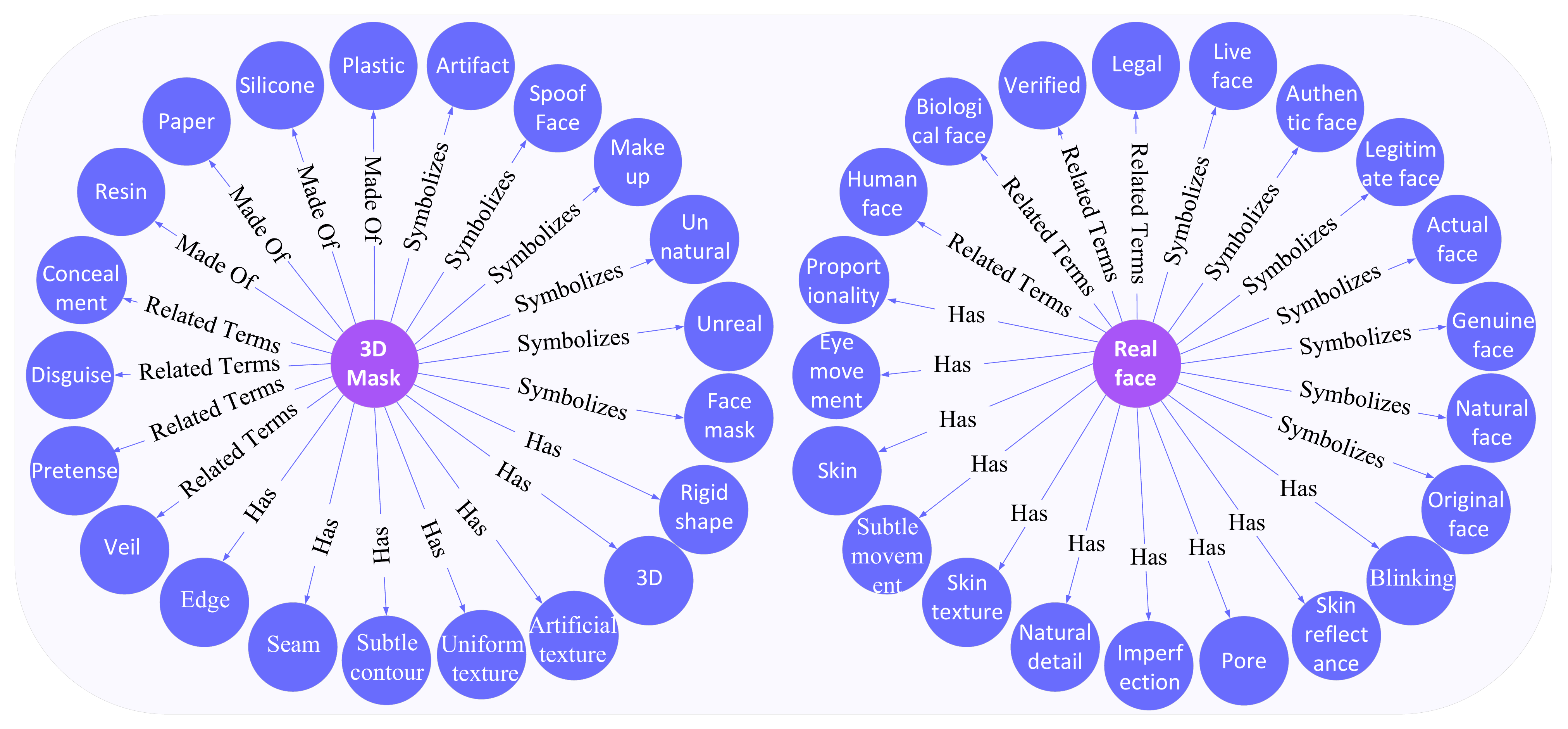}
		 \caption{\centering Overview of the knowledge graph constructed for 3D mask presentation attack detection.}
        %\makebox[\linewidth][c]{%
        %        \parbox{.9\linewidth}{\caption{Overview of the knowledge graph constructed for 3D mask presentation attack detection.}}%
        %     }
		\label{fig25}
	\end{figure*}

	The final knowledge graph constructed for 3D mask presentation attack detection, referred to as MaskPAD-KG, is illustrated in \figurename~\ref{fig25}. Formally, MaskPAD-KG is defined as $\mathcal{G=(E, R, S)}$, where $\mathcal{E} = \{e_1,e_2,\ldots,e_{|\mathcal{E}|}\}$ denotes the set of entities, and $\mathcal{R} = \{r_1,r_2,\ldots,r_{|\mathcal{R}|}\}$ represents the set of relationships. Entities and relationships are combined to form the set of triples $\mathcal{S}$. 
    Each triple $(h, r, o) \in \mathcal{S}$ consists of a head entity $h$, a tail entity $o$, and a relationship $r$ linking them. As illustrated in~\figurename~\ref{fig25},  MaskPAD-KG contains a total of 44 entities, 4 types of relationships, and 42 triples.
    Both the entity set $\mathcal{E}$ and the triple set $\mathcal{S}$ encapsulate domain-specific knowledge essential for 3D mask presentation attack detection. We encode this structured knowledge into prompts by leveraging both the entity set and the triple set.

	\begin{figure}[!t]
		\centering
		\includegraphics[width=0.5\textwidth]{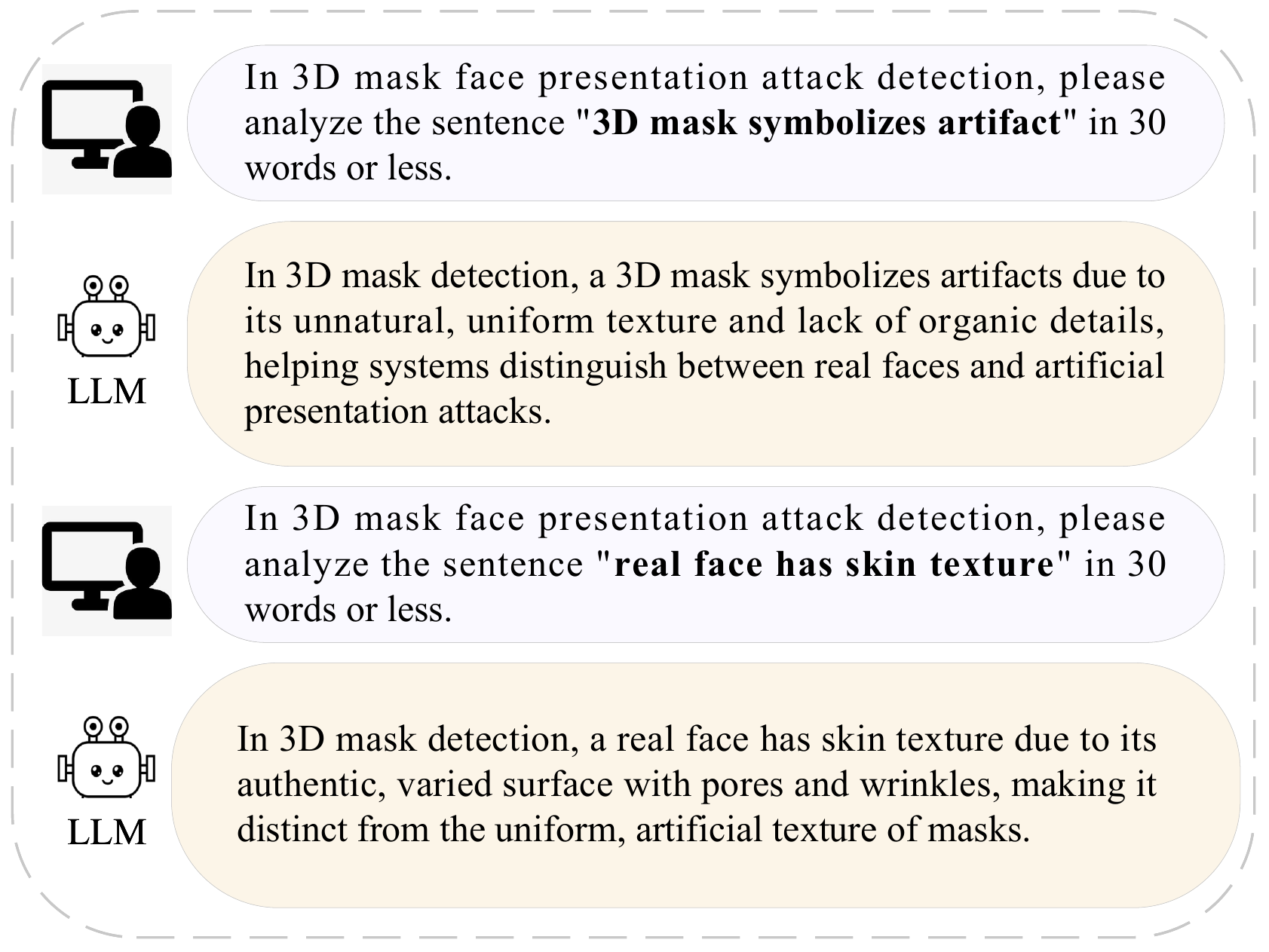}
		\caption{ Examples of fine-grained discriminative descriptions. Each quotation-marked phrase in the question illustrates an instance of a triple.}
		\label{fig17}
	\end{figure}

	\noindent \textbf{Knowledge graph prompt generation.}
	Given the $k$-th category name $c^k$ in $\mathcal{C}$ ($k \in \{1,2,\ldots,K\}$), we combine all category-related entities in $\mathcal{E}$ following the pattern shown in Equation~\ref{equ33} to generate the category-associated entity prompt $t^k_e$.
	\begin{equation}\label{equ33}
		\begin{aligned}
			t^k_e=[c^k][e_1^k][e_2^k]\ldots[e^k_{{L^k_e}}],
		\end{aligned}
	\end{equation} 
	where $L^k_e$ is the number of entities associated with the $k$-th class name.
	
	The entities in the knowledge graph represent structured domain knowledge, but some of them are relatively abstract (e.g., \textit{pore}) and lack task-specific contextual information. This results in an overly large search space for vision-language models, making it difficult to accurately extract the pre-trained knowledge relevant to the 3D mask presentation attack task.
	To provide richer contextual information pertinent to 3D mask presentation attack detection, we further transform the triples in the knowledge graph into the fine-grained discriminative descriptions to be used in conjunction with the entity prompts.
	
	Large language models (e.g., GPT-4) trained on vast datasets contain a wealth of knowledge.  Given the $k$-th category name $c^k$ in $\mathcal{C}$ ($k \in \{1, 2, \ldots, K\}$), we extract the knowledge from the large language model to automatically generate the description for every triple $(h, r, o) \in \mathcal{S}$ related to $c^k$. 
	The question template used to query the large language model is as follows:
	“In 3D mask face presentation attack detection, please analyze the sentence [$h$][$r$][$o$] in 30 words or less.”
	One example is illustrated in \figurename~\ref{fig17}. The fine-grained discriminative description $t^k_d$ for $c^k$ is formally defined as:
	\begin{equation}\label{equ8}
		\begin{aligned}
			t^k_d=[c^k][d_1^k][d_2^k]\ldots[d^k_{{L^k_d}}],
		\end{aligned}
	\end{equation} 
	where $d_i^k$ represents the description corresponding to the $i$-th triple in $\mathcal{S}$ associated with the $k$-th category name $c^k$, $L^k_d$ is the number of descriptions associated with the $k$-th category name.
	 
	These fine-grained descriptions provide discriminative information for the vision-language model to distinguish between real faces and 3D masks. Furthermore, such descriptions efficiently incorporate task-specific contextual information, complementing the knowledge graph entity prompts. Expanding the knowledge within these descriptions helps address the gaps caused by incomplete knowledge graph entities, ultimately enhancing the robustness of the designed prompts in the face of changes to the knowledge graph entities.
	
	%HERE
	\noindent \textbf{Visual-specific knowledge filter.} A single input image is typically associated with only a subset of the information contained in the knowledge graph. The relevance between different images and knowledge graph entities varies, and similar variations exist for the corresponding fine-grained discriminative descriptions. To achieve better alignment between visual and textual modalities, we introduce visual-specific knowledge filters, which dynamically select and weight relevant knowledge terms (both entities and descriptions) by learning a soft attention distribution over candidate prompts, conditioned on the visual content of the input image.
 
\begin{figure}[!t]
		\centering
		\includegraphics[width=0.45\textwidth]{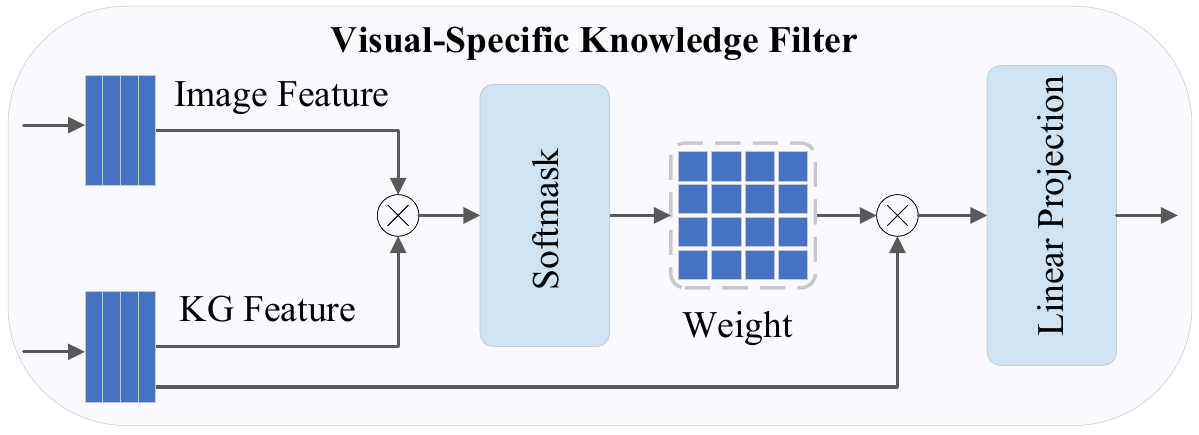}
		\caption{\centering Structure of the visual-specific knowledge filter. 
		}
		\label{fig18}
	\end{figure}

    Given the distinct alignment patterns between images and entities, as well as between images and descriptions, as illustrated in~\figurename~\ref{fig2}, we design two separate filtering modules: a visual-specific entity filter $\mathfrak{F}_e$ and a visual-specific description filter $\mathfrak{F}_d$. Each is responsible for selectively weighting its corresponding textual components based on their relevance to the input image.
    The visual-specific entity filter $\mathfrak{F}_e$ assigns higher weights to entities that are more semantically aligned with the visual input, while the visual-specific description filter $\mathfrak{F}_d$ performs a similar function for fine-grained discriminative descriptions. 
    Although these two filters operate on different types of knowledge graph features, they share an identical internal architecture, as shown in~\figurename~\ref{fig18}.

    To better adapt the image features for the 3D mask presentation attack detection task and enhance alignment with textual prompts, we employ an adaptation layer composed of two fully connected layers. This module transforms the original visual representation into a task-specific embedding, denoted as $f_v$.
    Let $f_{kg}(t)$ represent the token embedding of a textual prompt $t$ (either for entities or for descriptions), with $m$ denoting the dimensionality of its last dimension. A linear projection layer, denoted by $\varPsi$, is then applied to compute the output of a visual-specific knowledge filter $\mathfrak{F}$ for prompt $t$ as follows:
	\begin{equation}\label{equ9}
		\begin{aligned}
			\mathfrak{F}(t)=\varPsi(Softmax(\frac{f_v\cdot f_{kg}(t)^{\top}}{\sqrt{{m}}})\cdot f_{kg}(t)).
		\end{aligned}
	\end{equation}
    For a given category name $c^k$, we concatenate the entity prompt refined by $\mathfrak{F}_e$ and the description refined by $\mathfrak{F}_d$ to form its final visual-specific knowledge graph prompt $t_{vskg}^k$:	
	\begin{equation}\label{equ99}
		\begin{aligned}
			t_{vskg}^k=\mathfrak{F}_e(t_e^k)\oplus\mathfrak{F}_d(t_d^k).
		\end{aligned}
	\end{equation}
    The proposed visual-specific knowledge filters play a critical role in suppressing irrelevant or potentially misleading information that may exist in the raw knowledge graph or generated descriptions. By tailoring the selected knowledge to the visual input, the generated visual-specific knowledge graph prompt offers accurate and context-aware guidance, thereby enhancing both the robustness and generalization ability for 3D mask presentation attack detection.
	
	In addition to the visual-specific knowledge graph prompt, we also design the learnable context prompt. Given the $k$-th category name $c^k$ in $\mathcal{C}$ ($k \in \{1, 2, \ldots, K\}$), its prompt $t^k_l$ consists of the token embedding of the prefix context sequence and the category name:
	\begin{equation}\label{equ10}
		\begin{aligned}
			t^k_l=[v_1^k][v_2^k]\ldots[v^k_{L_l^k}][c^k],
		\end{aligned}
	\end{equation} 
	where $[v_i^k]$ ($i \in \{1,2,\ldots,L_l^k\}$) is the $i$-th context vector for $c^k$ and $L_l^k$ is the number of context tokens.
	Ultimately, the visual-specific knowledge graph prompt and the learnable context prompt are concatenated and fed into the text encoder to obtain the knowledge-based text feature $f_t^k$.
	
	\subsection{Spurious Correlation Elimination Prompt Learning}
	\label{33} 
	Compared to tasks like image recognition, which emphasize the overall structure and global appearance of objects, 3D mask presentation attack detection relies on identifying subtle visual cues, such as minute variations in color, texture, and structure.  However, visual-language models trained on general-purpose data may struggle to capture the classification patterns specific to the 3D mask presentation attack detection. These discriminative cues are also susceptible to scenario-specific factors such as recording environment, equipment, and lighting conditions, which pose significant challenges for 3D mask presentation attack detection. 
	Motivated by causal representation learning, we further impose constraints on the generated text features to explicitly eliminate correlations with scenario-specific patterns while preserving alignment with the critical features that distinguish real faces from 3D masks. This promotes the learning of causal invariance, allowing a more generalizable detection.

    \begin{figure}[!t]
		\centering
		\includegraphics[width=0.45\textwidth]{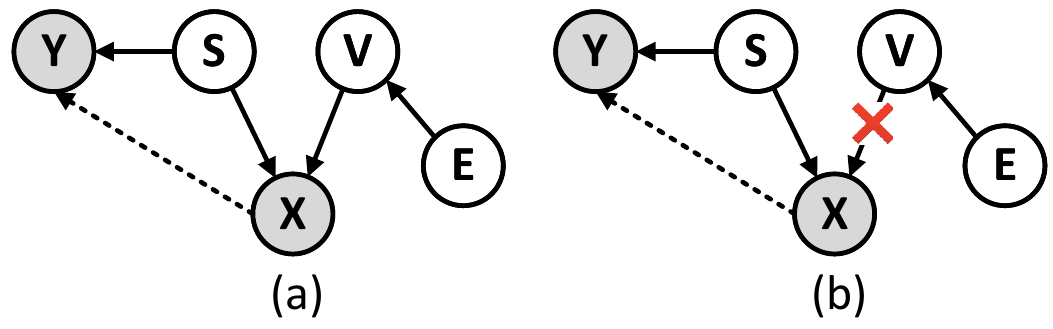}
		\caption{Causal graph for multiple elements in 3D mask presentation attack detection. Gray variables are directly observable, while white nodes are unobservable. 
		}
		\label{fig16}
    \end{figure}
    
	\noindent \textbf{Causal graph analysis.} 
    To provide a principled foundation for the constraint design, we construct a causal graph that explicitly models the dependency structure among critical variables influencing 3D mask presentation attack detection, as illustrated in \figurename~\ref{fig16}(a). Here, nodes $X$ and $Y$ represent the observable input image and its corresponding label, respectively; $E$ represents external confounders, while nodes $S$ and $V$ refer to unobservable factors relevant and irrelevant to 3D mask presentation attack detection. The detection-relevant factors $S$ capture critical attributes that differentiate real faces from 3D masks, such as material and facial structure. These attributes typically exhibit strong cross-scenario generalization capabilities.
    In contrast, the detection-irrelevant factors $V$ represent external, scenario-specific influences, such as the recording environment and equipment. While detection-irrelevant factors do not pertain to the classification of real faces versus 3D masks, they significantly affect the appearance of the image and vary across different scenarios. Both $S$ and $V$ jointly shape the observable presentation of the image $X$.
    
	The label $Y$ should be determined exclusively by the detection-relevant factors $S$. However, since $S$ is unobservable, models must rely on the relationship between $X$ and $Y$. This reliance creates vulnerability to domain shifts, as the widely used empirical risk minimization objective $P(Y|X)$ inherently conditions on $X$. Consequently, spurious correlations emerge between $V$ and $Y$ through the causal pathway $Y \leftarrow S \rightarrow X \leftarrow V$. As a result, the learned models often incorporate detection-irrelevant yet environment-specific information, impairing the generalization capability. To address this issue, it is essential to break the spurious causal path and eliminate the correlation between $V$ and $Y$, as illustrated in \figurename~\ref{fig16}(b). We propose enforcing a causal independence property, $Y \perp V$, which ensures that $Y$ is independent of $V$. This approach constrains the model to focus solely on detection-relevant features, excluding environment-specific information. By doing so, we enhance the robustness and generalization capability of models.
	
	\begin{figure}[!t]
		\centering
		\includegraphics[width=0.49\textwidth]{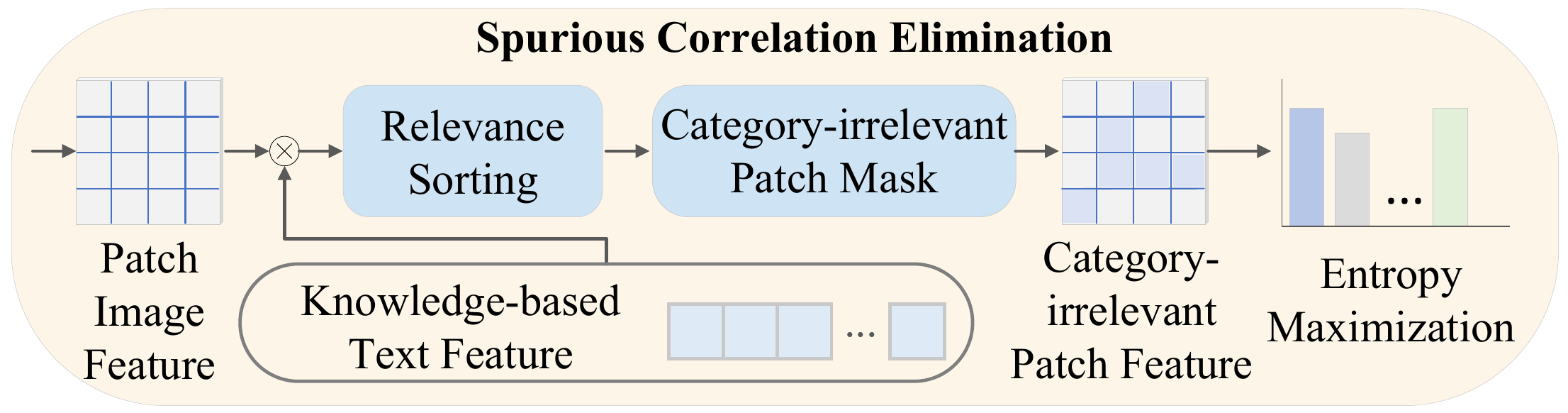}
		\caption{Visualization of spurious correlation elimination regularization. 
		}
		\label{fig19}
	\end{figure}
	
	\noindent \textbf{Spurious correlation elimination regularization.} In this work, we represent the label $Y$ using category-related prompts. To enforce the causal independence property $Y \perp V$, we introduce a spurious correlation elimination regularization that encourages the generated prompts to be independent of detection-irrelevant factors $V$. An overview of this regularization strategy is presented in~\figurename~\ref{fig19}. Prior studies have shown that local fine-grained visual information can offer effective discriminative cues for 3D mask presentation attack detection~\cite{akhtar2016face,atoum2017face}. Accordingly, our approach aims to suppress spurious correlations between generated prompts and category-irrelevant local patches within the input image.
	
	 To this end, we first divide each input image into non-overlapping local patches. To assess the relevance of each patch for a sample $(x_i, y_i)$, we compute its similarity with the prompts of all categories. The similarity $s_j^k$ between the $j$-th patch and the prompt of the $k$-th category is defined as:
	\begin{equation}\label{equ3}
		\begin{aligned}
			s_j^k=\frac{\exp\left(\cos\left(f_{v}^{j},f_t^k\right)/\tau\right)}{\sum_{k^{\prime}=1}^K\exp\left(\cos\left(f_{v}^{j},f_t^{k^{\prime}}\right)/\tau\right)},
		\end{aligned}
	\end{equation}
	where $f_{v}^{j}$ is the local feature of the $j$-th patch, $f_t^k$ denotes the knowledge-based text feature of the $k$-th category, and $\tau$ is the temperature coefficient.
	
	A patch is deemed category-irrelevant if its most similar category does not correspond to the ground-truth label $y_i$. Thus, the set of category-irrelevant patches $\Omega$ for the sample $(x_i, y_i)$ is defined as:
	\begin{equation}\label{equ4}
		\begin{aligned}
			\Omega=\{j| \underset{1\leq k \leq K}{{\arg\max} \, }(s_j^k)\neq y_i  \}. 
		\end{aligned}
	\end{equation}
	To mitigate spurious correlations between the generated prompts and the identified category-irrelevant patches, we maximize the entropy of their similarity distribution $s_j=\{s_j^k \mid k \in \{1,2,\ldots,K\}\}$ via the following spurious correlation elimination regularization:
	\begin{equation}\label{equ5}
		\begin{aligned}
			\ell_{sce}\left(x_i,y_i\right)=\sum_{j\in\Omega}\sum_{k=1}^\mathrm{K}s_j^k\log s_j^k. 
		\end{aligned}
	\end{equation}  
	Here, $s_j^k$ can be interpreted as the probability of the $j$-th patch belonging to the $k$-th category. Maximizing the entropy enforces a uniform distribution across $K$ categories, thereby reducing the chance that generated prompts spuriously align with irrelevant patches. This promotes the independence of generated prompts from category-irrelevant content and mitigates the influence of detection-irrelevant factors.
	
	\noindent \textbf{Spoof-relevant discriminant regularization.} 
	Beyond eliminating spurious correlations, it is equally important to strengthen the correlations between detection-relevant factors and ground-truth labels. To this end, we introduce a spoof-relevant discriminant regularization $\ell_{srd}$, which aligns image features with the knowledge-based text features of the ground-truth category.
	Formally, $\ell_{srd}$ is defined as a cross-entropy loss:
	\begin{equation}\label{equ6}
		\begin{aligned}
			\ell_{srd}\left(x_i,y_i\right)=-\sum_{k=1}^{K} y^k \log d^k.
		\end{aligned}
	\end{equation}
	Here, $y^k$ is a binary indicator such that $y^k=1$ if $y_i$ is the $k$-th category in the label set $\mathcal{C}$, and $y^k=0$ otherwise. $d^k$ is the probability assigning the input image $x_i$ to the $k$-th category, which is computed as: 
	\begin{equation}\label{equ7}
		\begin{aligned}
			d^k=\frac{\exp\left(\cos\left(f_{v},f_t^k\right)/\tau\right)}{\sum_{k^{\prime}=1}^K\exp\left(\cos\left(f_{v},f_t^{k^{\prime}}\right)/\tau\right)}.
		\end{aligned}
	\end{equation}
	The value of $d^k$ reflects the similarity between the image feature $f_v$ of the input $x_i$ and the text feature $f_t^k$ associated with the $k$-th category, and can be interpreted as the predicted probability for that category.
	
	This regularization encourages the model to produce a peaked probability distribution over the $K$ categories, with the highest probability assigned to the ground-truth class. It promotes maximum alignment between the image features and the text features of the corresponding category, thus reinforcing the model’s  focus on detection-relevant factors. Consequently, the spoof-relevant discriminant regularization enhances both the discriminative capacity and generalization ability of the model across diverse scenarios.
	
	\subsection{Overall Objective and Inference}
	\label{34}
	The overall training objective $\ell_{all}$ combines the two proposed regularizations and is defined as:
	\begin{equation}\label{equ21}
		\ell_{all} = \ell_{srd} + \lambda \ell_{sce},
	\end{equation}
	where $\lambda$ is a hyperparameter that balances the contribution of the two regularizations.
	
	During inference, we compute the predicted probability distribution of a given sample based on Equation~\ref{equ7}.
	Specifically, the trained model is first used to compute $d^k$ for each category according to Equation~\ref{equ7}, obtaining the similarity scores between the image feature and the knowledge-based text feature of the $k$-th category. All the $K$ similarity scores are aggregated to form a set as follows:
	\begin{equation}\label{equ18}
		d = \{d^k \mid k \in \{1,2,\ldots,K\}\}.
	\end{equation}
	This set of similarity scores serves as the predicted probability distribution over the label set. A threshold-based decision rule is subsequently applied to classify the test sample as either a real face or a 3D mask. If the predicted probability of a test sample being a real face exceeds the threshold, the sample is classified as a real face; otherwise, it is classified as a 3D mask.
	The decision threshold is determined on the development set and corresponds to the point at which the False Rejection Rate (FRR) equals the False Acceptance Rate (FAR), ensuring a balanced trade-off between the two types of classification errors.
	
	\section{Experiments}
	\label{sec4}
     This section presents comprehensive experimental evaluations of the proposed method. We begin by describing the experimental setup, including datasets, protocols, and implementation details. Next, we compare the proposed approach with existing state-of-the-art methods under both intra-dataset and cross-dataset evaluation settings. Finally, we perform extensive ablation studies and provide qualitative visualizations to analyze the effectiveness of each component within our framework.
	\subsection{Experimental Setups}
	\label{sec41}
	\noindent \textbf{Datasets.}
	We evaluate the performance of the proposed method on three widely used datasets: 3DMAD (3D Mask Attack
	Dataset)~\cite{erdogmus2014spoofing}, HKBU-MARs V1+ (Hong Kong Baptist University 3D Mask Attack with Real-World Variations)~\cite{liu2018remote} and HiFiMask (High-Fidelity Mask dataset)~\cite{liu2022contrastive}. 
	%\figurename~\ref{fig4} shows examples from three datasets.
	Note that only RGB images are used in our experiments.
	
	The 3DMAD dataset consists of 255 videos containing both RGB and depth images captured using a Kinect camera from 17 subjects. The real facial data is recorded in two distinct controlled scenarios, while the mask data is collected in a different scenario. The masks, customized by ThatsMyFace.com, are crafted from hard resin based on the real facial data, accurately depicting facial details such as beards and moles. %These masks feature openings for the eyes and nostrils that align with the real faces, although the mouths of the masks remain fixed.
	
	The HKBU-MARs V1+ dataset includes 180 videos of 12 subjects recorded in an office environment using a Logitech C920 camera. In contrast to the 3DMAD dataset, the HKBU-MARs V1+ dataset includes 3D masks custom-made by both ThatsMyFace.com and REAL-f.jp. Masks created by REAL-f.jp exhibit better alignment with real faces regarding color, texture, and three-dimensional structure than those made by ThatsMyFace.com.
	
	The HiFiMask dataset offers a more extensive and diverse collection of data volume, recorded subjects, recording scenarios, and capturing devices. It comprises 54,600 videos of 75 subjects. Masks in this dataset are crafted from three materials: transparent, plaster, and resin. Videos are recorded across six scenarios, with each scenario including six distinct lighting conditions. Seven different image-capturing devices equipped with high-resolution cameras are used for these recordings.
	%\begin{figure}[!t]
	%	\centering
	%	\includegraphics[width=0.48\textwidth]{fig4.pdf}
	%	\caption{Examples from the evaluated datasets. The first row in each dataset is real faces, while the second row is 3D masks. These datasets exhibit significant differences in mask material, shape, recording background, lighting, and device.
	%	}
	%	\label{fig4}
	%\end{figure}
	
	\begin{table*}[!tbp]
		%\scriptsize
		\rowcolors{1}{}{lightgray}
		\footnotesize
		\caption{Intra-dataset evaluation results on the HKBU-MARs V1+ and  HiFiMask datasets. ‘U’: unavailable.}
		\label{tabintra1}
		\centering 
		\begin{tabular}{p{2cm}p{1.5cm}<{\centering}p{0.85cm}<{\centering}p{0.85cm}<{\centering}p{1.2cm}<{\centering}p{1.3cm}<{\centering}p{1.15cm}<{\centering}p{0.85cm}<{\centering}p{0.85cm}<{\centering}p{1.2cm}<{\centering}p{1.3cm}<{\centering}
			}
			\hline
			Method& \multicolumn{5}{c}{HKBU-MARs V1+ } & \multicolumn{5}{c}{ HiFiMask}  \\
			%\hline
			& HTER$\downarrow$&EER$\downarrow$ &AUC$\uparrow$ &B@A=0.1$\downarrow$
			&B@A=0.01$\downarrow$& HTER$\downarrow$&EER$\downarrow$ &AUC$\uparrow$ &B@A=0.1$\downarrow$
			&B@A=0.01$\downarrow$  \\
			%\hline
			MS-LBP~\cite{maatta2011face}&24.00$\pm$25.60& 22.50 &85.80& 48.60 &95.10&48.30&40.50 &63.70 &78.90& 97.30\\ 
			%\hline
			CTA ~\cite{boulkenafet2016face}&23.40$\pm$20.50&23.00& 82.30 &53.80 &89.20&40.70 &31.60& 74.90 &64.10 &92.90\\ 
			%\hline
			CDCN++~\cite{yu2020searching} & 4.83$\pm$7.60& 8.70& 96.00 &7.77 &66.20&3.67&2.64& 99.60 &0.83 &6.79\\ 
			%\hline
			HRFP~\cite{grinchuk20213d}  &3.34$\pm$5.70& 4.33& 99.20&1.34&6.23&2.20&2.26& 99.70 &0.28& 4.35\\ 
			%\hline
			ViTranZFAS~\cite{george2020effectiveness} &U& U& U& U& U& 2.63&2.48& 99.70&0.21&6.58\\ 
			%\hline
			MD-FAS~\cite{guo2022multi} &U& U& U& U& U& 5.45&4.12& 99.40& 0.34& 15.20\\ 
			%\hline
			CFrPPG~\cite{liu2021multi} &42.10$\pm$5.60 &42.00 &60.80& U&U&U& U& U& U& U\\ 
			%\hline
			LeTSrPPG~\cite{liu2022learning} &15.80$\pm$6.50& 15.70 &91.50& U&U&U& U& U& U& U\\ 
			%\hline
			FASTEN~\cite{cao2024flow} &2.13$\pm$5.10 &2.56 &99.70 &0.81 &2.83&2.11& 2.18 &99.80 &\textbf{0.13} &4.13 \\ 
			%\hline 
			Ours&\textbf{1.65$\pm$3.81}&\textbf{0.73}&\textbf{99.97}&\textbf{0.03}&\textbf{0.67}&\textbf{1.52}&\textbf{1.38}&\textbf{99.83}&0.17&\textbf{2.22}\\ 
			\hline
		\end{tabular}
	\end{table*}
	\begin{table}[!tbp]
		%\scriptsize
		\footnotesize
		\rowcolors{1}{}{lightgray}
		\caption{Intra-dataset evaluation results on the 3DMAD dataset. ‘U’: unavailable.}
		\label{tabintra2}
		\centering \begin{tabular}{p{2cm}p{1.15cm}<{\centering}p{0.6cm}<{\centering}p{0.6cm}<{\centering}p{1.05cm}<{\centering}p{1.2cm}<{\centering}}
			\hline
			Method& \multicolumn{5}{c}{3DMAD }   \\
			%\hline
			& HTER$\downarrow$&EER$\downarrow$ &AUC$\uparrow$ &B@A=0.1$\downarrow$
			&B@A=0.01$\downarrow$ \\
			%\hline
			MS-LBP~\cite{maatta2011face}& 1.92$\pm$3.40& 3.28& 99.40& 0.32& 6.78\\ 
			%\hline
			CTA ~\cite{boulkenafet2016face}& 4.40$\pm$9.70& 4.24& 99.30& 1.60& 11.80\\ 
			%\hline
			CDCN++~\cite{yu2020searching} &4.20$\pm$7.10& 8.34 &96.70& 6.62 &59.60 \\ 
			%\hline
			HRFP~\cite{grinchuk20213d}  & 2.34$\pm$5.20 &2.41& 99.70& 0.67&5.87\\ 
			%			\hline
			%			ViTranZFAS~\cite{george2020effectiveness} &N/A& N/A& N/A& N/A& N/A\\ 
			%			\hline
			%			MD-FAS~\cite{guo2022multi} &N/A& N/A& N/A& N/A& N/A\\ 
			%\hline
			CFrPPG~\cite{liu2021multi} &32.70$\pm$7.40& 32.50& 70.80&U&U\\ 
			%\hline
			LeTSrPPG~\cite{liu2022learning} & 11.80$\pm$8.60& 11.90& 94.40&U&U\\ 
			%\hline
			FASTEN~\cite{cao2024flow} &1.17$\pm$0.70& 1.06 &99.80 &0.03 &0.31\\ 
			%\hline
			Ours&\textbf{0.33$\pm$0.56}&\textbf{0.11}&\textbf{100.00}&\textbf{0.00}&\textbf{0.00}\\ 
			\hline
		\end{tabular}
		
	\end{table}
	
	\begin{table*}[!htbp]
		%\scriptsize
		\footnotesize
		\rowcolors{1}{}{lightgray}
		\caption{Cross-dataset evaluation results under the 3DMAD$\rightarrow$HKBU-MARs V1+ and HKBU-MARs V1+ $\rightarrow$3DMAD protocols. ‘U’: unavailable.}
		\label{tabcro1}
		\centering \begin{tabular}{p{2cm}p{1.5cm}<{\centering}p{0.85cm}<{\centering}p{0.85cm}<{\centering}p{1.2cm}<{\centering}p{1.3cm}<{\centering}p{1.15cm}<{\centering}p{0.85cm}<{\centering}p{0.85cm}<{\centering}p{1.2cm}<{\centering}p{1.3cm}<{\centering}
			}
			\hline
			Method& \multicolumn{5}{c}{3DMAD$\rightarrow$HKBU-MARs V1+} & \multicolumn{5}{c}{HKBU-MARs V1+ $\rightarrow$3DMAD}  \\
			%\hline
			& HTER$\downarrow$&EER$\downarrow$ &AUC$\uparrow$ &B@A=0.1$\downarrow$
			&B@A=0.01$\downarrow$& HTER$\downarrow$&EER$\downarrow$ &AUC$\uparrow$ &B@A=0.1$\downarrow$
			&B@A=0.01$\downarrow$  \\
			%\hline
			MS-LBP~\cite{maatta2011face}&47.70$\pm$7.00 &48.30& 52.40& 86.40& 97.60& 43.20$\pm$7.30& 43.70& 58.80& 87.50& 99.20\\ 
			%\hline
			CTA ~\cite{boulkenafet2016face}&51.50$\pm$2.40& 55.30& 48.90& 90.50& 98.80& 68.20$\pm$7.70& 65.40& 40.10& 94.70& 99.90\\ 
			%\hline
			CDCN++~\cite{yu2020searching} & 50.30$\pm$2.70& 55.30& 42.30& 93.20& 99.90& 41.10$\pm$6.80& 33.10& 66.20& 74.10& 98.90\\ 
			%\hline
			HRFP~\cite{grinchuk20213d}  &25.80$\pm$3.10 &32.30& 69.20& 81.30& 98.40& 6.76$\pm$0.90& 7.25& 98.60& 5.63& 22.00\\ 
			%\hline
			%			ViTranZFAS~\cite{george2020effectiveness} &N/A& N/A& N/A& N/A&N/A&N/A&N/A& N/A& N/A&N/A\\ 
			%			\hline
			%			MD-FAS~\cite{guo2022multi} &N/A &N/A& N/A& N/A&N/A&N/A&N/A& N/A& N/A&N/A\\ 
			%			\hline
			CFrPPG~\cite{liu2021multi} &39.20$\pm$1.40& 40.10& 63.60& 75.50&U& 40.10$\pm$2.30& 40.60& 62.30& 79.10&U\\ 
			%\hline
			LeTSrPPG~\cite{liu2022learning} &15.70$\pm$0.50& 16.60 &90.10&25.20& U &12.90$\pm$0.80& 13.10 &93.40&15.80&U\\ 
			%\hline
			FASTEN~\cite{cao2024flow}& 11.80$\pm$2.10& 16.80& 91.00& 18.80& 24.40&
			3.85$\pm$1.90& 4.35& 99.10& 3.00 &8.88\\
			%\hline
			Ours
			&\textbf{2.32$\pm$6.09}&\textbf{0.13}&\textbf{97.96}&\textbf{1.19}&\textbf{1.21}
			&\textbf{2.13$\pm$2.14}&\textbf{0.01}&\textbf{99.85}&\textbf{0.33}&\textbf{0.54}
			\\ 
			\hline
		\end{tabular}
	\end{table*}
	
	\begin{table*}[!htbp]
		%\scriptsize
		\footnotesize
		\rowcolors{1}{}{lightgray}
		\caption{Cross-dataset evaluation results under the HiFiMask$\rightarrow$3DMAD and HiFiMask$\rightarrow$HKBU-MARs V1+ protocols.% ‘U’: unavailable. ‘N/A’: not applicable.
		}
		\label{tabcro2}
		\centering \begin{tabular}{p{2cm}p{1.5cm}<{\centering}p{0.85cm}<{\centering}p{0.85cm}<{\centering}p{1.2cm}<{\centering}p{1.3cm}<{\centering}p{1.15cm}<{\centering}p{0.85cm}<{\centering}p{0.85cm}<{\centering}p{1.2cm}<{\centering}p{1.3cm}<{\centering}
			}
			\hline
			Method& \multicolumn{5}{c}{HiFiMask$\rightarrow$3DMAD} & \multicolumn{5}{c}{HiFiMask$\rightarrow$HKBU-MARs V1+}  \\
			%\hline
			& HTER$\downarrow$&EER$\downarrow$ &AUC$\uparrow$ &B@A=0.1$\downarrow$
			&B@A=0.01$\downarrow$& HTER$\downarrow$&EER$\downarrow$&AUC$\uparrow$ &B@A=0.1$\downarrow$
			&B@A=0.01$\downarrow$  \\
			%\hline
			MS-LBP~\cite{maatta2011face}&47.80& 41.30& 62.60& 76.30& 98.00&50.40 &57.00& 39.10& 93.00& 98.80\\ 
			%\hline
			CTA ~\cite{boulkenafet2016face}&50.10 &71.50& 32.20& 97.60& 99.80&49.50 &63.30 &42.90& 90.00& 94.50\\ 
			%\hline
			CDCN++~\cite{yu2020searching} &36.80 &37.70& 72.10& 65.90 &87.10&35.70& 29.70& 80.80& 52.90 &78.40 \\ 
			%\hline
			HRFP~\cite{grinchuk20213d}  &21.30& 12.40& 95.80 &14.60&32.50& 7.68 &8.33& 99.10& 1.52& 11.40\\ 
			%\hline
			ViTranZFAS~\cite{george2020effectiveness} &26.30&16.50& 91.10& 23.10&60.80&32.70& 20.60& 88.30& 34.20&
			60.2\\ 
			%\hline
			MD-FAS~\cite{guo2022multi} &33.50& 34.40& 69.90& 67.00& 75.30&9.38& 8.61 &97.40 &5.67&
			43.40\\ 
			%\hline
			%CFrPPG~\cite{liu2021multi} &N/A &N/A& N/A& N/A&N/A&N/A&N/A& N/A& N/A&N/A\\ 
			%\hline
			%LeTSrPPG~\cite{liu2022learning} &N/A &N/A& N/A& N/A&N/A&N/A&N/A& N/A& N/A&N/A\\ 
			%\hline
			FASTEN~\cite{cao2024flow}  &2.35& 2.71 &99.80& 0.27 &4.76 &7.38 &8.55 &98.30 &5.40 &17.70\\
			%\hline
			Ours&\textbf{1.04}&\textbf{1.76}&\textbf{99.97}&\textbf{0.05}&\textbf{0.56}&
			\textbf{0.41}&\textbf{1.36}&\textbf{100.00}&\textbf{0.00}&\textbf{0.00}
			\\ 
			\hline
		\end{tabular}
	\end{table*}
	
	\noindent \textbf{Evaluation protocols and metrics.}
	We follow the evaluation protocols defined in~\cite{liu2021multi,cao2024flow} to evaluate the performance of the proposed method.
	The protocols include both intra-dataset and cross-dataset evaluations. To mitigate the impact of subject randomness, the performance of each protocol is averaged over 20 rounds.
	
	In the intra-dataset protocols, models are trained and tested using a single dataset. 
	Due to the limited number of subjects in the 3DMAD and HKBU-MARs V1+ datasets, we employ the leave-one-out cross-validation (LOOCV) protocol. One subject is randomly selected as the test set. For the 3DMAD dataset, eight subjects are randomly chosen as the training set, and the remaining eight subjects form the development set. For the HKBU-MARs V1+ dataset, five subjects are allocated to the training set, and the remaining six subjects form the development set.
	For the HiFiMask dataset, intra-dataset performance is evaluated using the officially provided Protocol 1.
	In the cross-dataset protocols, the training and test datasets are different.
	The protocols are constructed  as follows: 3DMAD $\rightarrow$ HKBU-MARs V1+, HKBU-MARs V1+ $\rightarrow$ 3DMAD, HiFiMask$\rightarrow$ 3DMAD, and  HiFiMask  $\rightarrow$ HKBU-MARs V1+. %Given the significantly richer data quantity and diversity of the HiFiMask dataset compared to the 3DMAD and HKBU-MARs V1+ datasets, we do not establish a protocol where HiFiMask serves as the test dataset.
	
	%\textbf{Evaluation metrics.} 
	We use the widely-used metrics to evaluate the proposed method against the state-of-the-art approaches, including Half Total Error Rate (HTER), Equal Error Rate (EER), Area Under Curve (AUC)~\cite{chingovska2012effectiveness}, Attack Presentation Classification Error Rate (APCER), and Bonafide Presentation Classification Error Rate (BPCER)~\cite{evabz}. 
	Here, ``B@A=0.1/0.01” represents the BPCER when APCER equals 0.1 or 0.01. Note that EER is calculated based on the development set, which shares a similar data distribution with the training set. In contrast,  HTER is computed using thresholds derived from the development set but applied to a test set with significantly different data distributions. Hence, EER reflects the model’s ability to fit the training data, while HTER measures its generalization capability on unseen test data.

	\noindent \textbf{Competitors.}
	We select nine representative face presentation attack detection approaches for performance comparison. They are listed as follows. 1) MS-LBP~\cite{maatta2011face} and CTA~\cite{boulkenafet2016face}: Classical methods that extract color and texture features using handcrafted feature descriptors. 2) CDCN++~\cite{yu2020searching}: A widely used approach that leverages central difference convolutional neural networks to learn fine-grained features. 3) HRFP~\cite{grinchuk20213d}: A method that extracts fine-grained information from local image patches and is the winner of the 3D High-Fidelity Mask Attack Detection Challenge at ICCV 2021. 4) ViTranZFAS~\cite{george2020effectiveness} and MD-FAS~\cite{guo2022multi}: Approaches that utilize pre-trained models to differentiate between real and fake faces. 5) CFrPPG~\cite{liu2021multi} and LeTSrPPG~\cite{liu2022learning}: Techniques that learn remote photoplethysmography (rPPG) features, demonstrating strong discriminative capability for 3D mask presentation attack detection. 6) FASTEN~\cite{cao2024flow}: A recent state-of-the-art method that learns spatiotemporal features, achieving superior performance on commonly used datasets.
	To ensure fairness, we follow the settings of the original papers and use the performance data reported in those studies for comparison. If the original papers do not report performance data for certain protocols, we mark them with "U", indicating that the data is unavailable.
	
	\noindent \textbf{Implementation details.}
	We use the face detector~\cite{Zhang2016Joint} to detect and align the face images from the original video frames.
	The proposed method is implemented in PyTorch, utilizing the CLIP model with a ViT-B/16 backbone as the pre-trained vision-language model. Aligned images are resized to $224 \times 224$ pixels before being passed to the image encoder. The number of classes $K$ is set to 2, and the number of context tokens $L_l^k$ is also set to 2.
	To optimize the generated prompts, the stochastic gradient descent (SGD) scheme, with a momentum of 0.9 and a weight decay of 0.0005, is used. 
	The batch size is set to 128, with the learning rate initialized at 0.001, and adjusted during training using cosine annealing.
	The scale factor $\lambda$ is set to 0.5.
	
	\subsection{Comparison with State-of-the-Art Methods}
 	In this subsection, we compare and analyze our method against existing state-of-the-art methods under both intra-dataset and cross-dataset protocols.
	\subsubsection{Intra-Dataset Evaluation}
	We first perform intra-dataset testing to evaluate the detection performance.
	The experimental results are shown in Table~\ref{tabintra1} and Table~\ref{tabintra2}. 
	The methods based on handcrafted LBP features~\cite{maatta2011face,boulkenafet2016face} achieve good performance on the 3DMAD dataset. 
	These results can be attributed to that the 3DMAD dataset is relatively simpler in terms of mask types, recording scenarios, and capturing devices.
	However, these methods are prone to overfitting to the patterns of the training set and thus exhibit poor performance on the more complex HKBU-MARs V1+ and HiFiMask datasets.
	In contrast, the performance of deep learning-based methods~\cite{yu2020searching,grinchuk20213d,george2020effectiveness,guo2022multi} is relatively consistent across all three datasets. 
	The rPPG-based methods~\cite{liu2021multi,liu2022learning} have significant room for improvement due to their sensitivity to noise interference. FASTEN~\cite{cao2024flow}, which aggregates spatio-temporal features, demonstrates strong performance on all three datasets.
	
	Compared to FASTEN, the proposed method achieves HTER improvements of 71.79\%, 22.54\%, and 27.96\% on the 3DMAD, HKBU-MARs V1+, and HiFiMask datasets, respectively.
	The proposed method also achieves significant performance gains across the other four evaluation metrics: EER, AUC, B@A=0.1, and B@A=0.01.
	Notably, while FASTEN requires multiple frames as input for 3D mask presentation attack detection, the proposed method requires only one single frame as input, yet achieves significant performance gains.
	These results demonstrate that integrating knowledge graphs and causal prompt learning effectively  adapts pre-trained vision-language models for 3D mask presentation attack detection.
	
	\subsubsection{Cross-Dataset Evaluation}
	\textbf{Quantitative comparison.} Significant distribution discrepancies exist among the three datasets due to differences in mask types, spoof face generation processes, recording backgrounds, illumination conditions, and recording devices. Cross-dataset protocols are used to evaluate the generalization ability of various methods against these factors. The evaluation results are presented in Table~\ref{tabcro1} and Table~\ref{tabcro2}.
	
	As shown in the results, the methods based on handcrafted LBP features~\cite{maatta2011face,boulkenafet2016face}  exhibit poor performance under cross-dataset protocols, demonstrating their limited generalization ability. 
	Similarly, unseen factors across different scenarios present substantial challenges for traditional deep learning-based methods~\cite{yu2020searching,grinchuk20213d,george2020effectiveness,guo2022multi}, which show a notable decline in performance compared to intra-dataset evaluations. By contrast, the rPPG-based methods~\cite{liu2021multi,liu2022learning} demonstrate minimal performance differences between intra-dataset and cross-dataset protocols, underscoring the strong generalization capability of heart rate features.
	
	The proposed method achieves HTER improvements of 80.33\%, 44.68\%, 55.74\%, and 94.44\% across the four protocols compared to FASTEN. 
	The performance improvements on the other four evaluation metrics are also significant. The significant performance gains can be attributed to the integration of fine-grained expert knowledge into the prompts of the vision-language model, enabling effective adaptation to the 3D mask presentation attack detection task.
	The HKBU-MARs V1+ dataset includes masks with higher realism compared to the 3DMAD dataset. 
	The substantial performance improvement under the 3DMAD$\rightarrow$HKBU-MARs V1+ protocol demonstrates the strong generalization ability of the proposed method to unseen mask types with unfamiliar colors, textures, materials, and structural patterns. 
	Additionally, the smaller performance gap at B@A=0.1 and B@A=0.01 also underscores the robustness of the proposed method to threshold variations.
	
\begin{figure}[!t]
		\centering
		\includegraphics[width=0.48\textwidth]{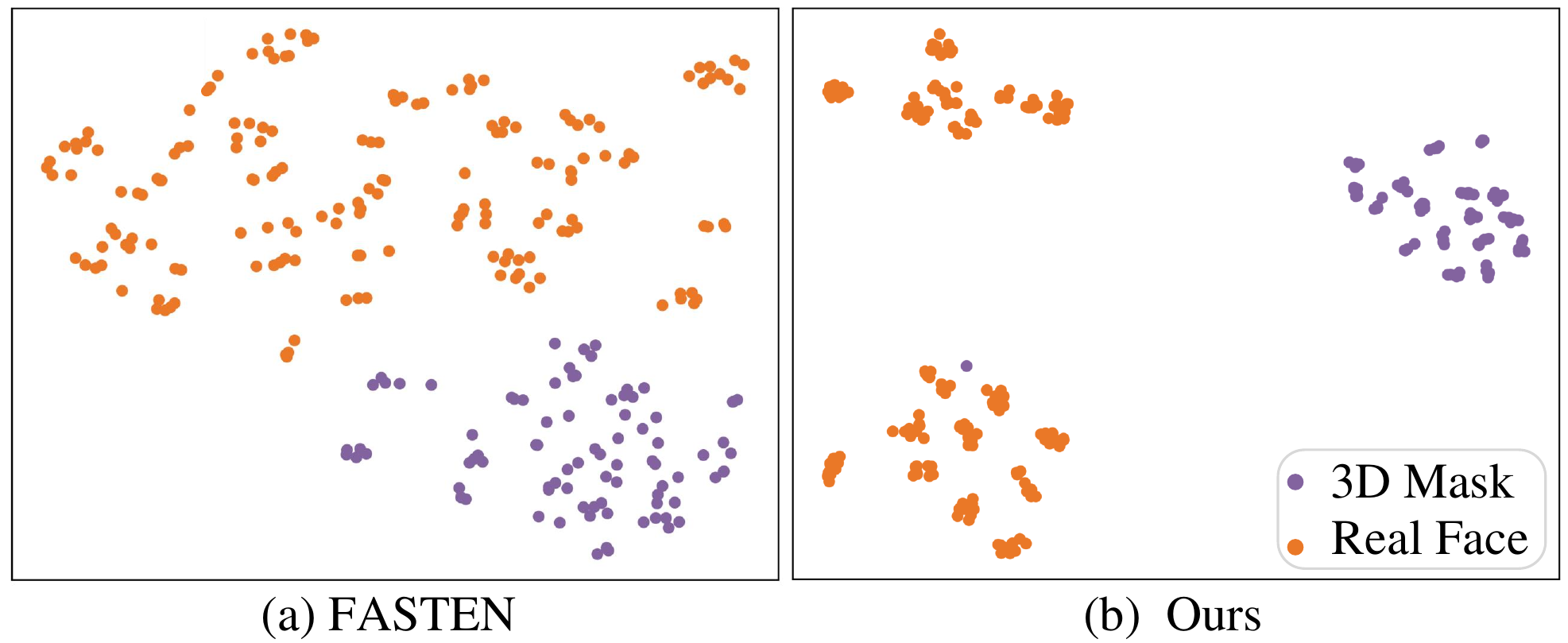}
		\caption{t-SNE plot of test domain features for (a) FASTEN and (b) our method under the HiFiMask$\rightarrow$3DMAD protocol.
		}
		\label{fig5}
	\end{figure}

	\noindent \textbf{Comparison of feature visualizations.}
	  We employ t-SNE to visualize the image feature distributions of the test domain, as extracted by FASTEN and our method under the HiFiMask$\rightarrow$3DMAD protocol, as shown in \figurename~\ref{fig5}. Since FASTEN utilizes multi-frame temporal information, each point in \figurename~\ref{fig5} corresponds to one video sample.
	It can be observed that the sample features extracted by FASTEN exhibit dispersed intra-class distributions and narrow inter-class margins. In contrast, our method produces sample features with compact intra-class distributions and significantly wider inter-class separations.
	Since the test domain is not accessible during training, the distribution of features indicates the generalization capability of the learned models.
	Although the visualization of FASTEN suggests that the test samples appear to be linearly separable, the dispersed feature layout and narrow inter-class margins imply that the decision boundary derived from the training data may not perfectly traverse this narrow margin, potentially leading to misclassifications.
	In contrast, the wider inter-class margins and tighter intra-class clustering produced by our method enable the decision boundary to generalize more robustly to the test domain, as it is more likely to fall within the expanded inter-class margin.
	Notably, as show in \figurename~\ref{fig5}(b), our method aggregates real faces into two clusters. In the 3DMAD dataset, the real face data originate from two distinct scenarios. Our approach tends to map samples with similar category-irrelevant factors together.
	
	\begin{figure}[!htbp]
	\centering
	\includegraphics[width=0.4\textwidth]{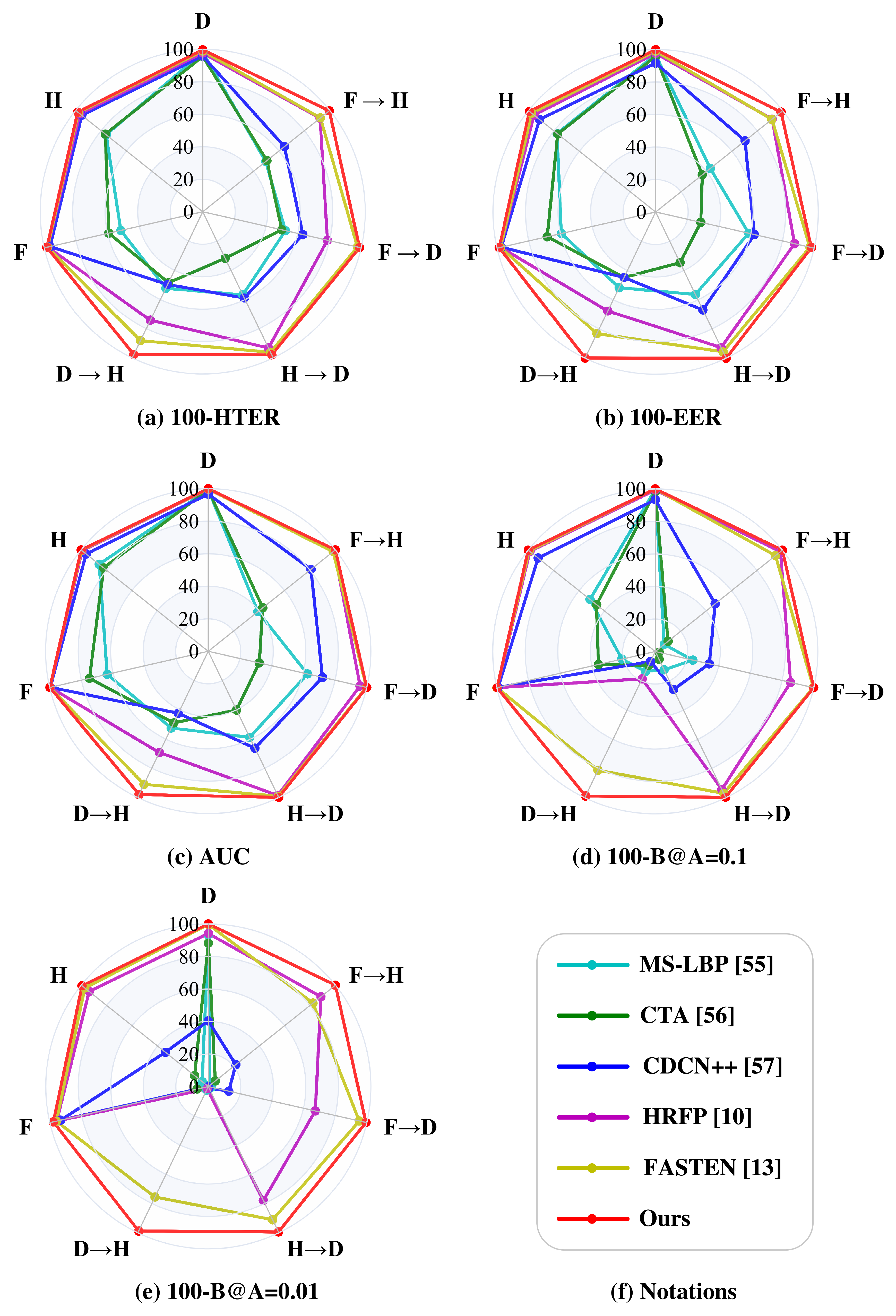}
	\caption{Overall performance comparison across all protocols.
	}
	\label{fig6}
	\end{figure}

	\subsubsection{Overall Performance Evaluation}
	We employ a radar chart to visualize the performance of the proposed method compared to existing methods across the seven protocols, providing a comprehensive evaluation of overall performance, as shown in \figurename~\ref{fig6}. For clarity, the datasets 3DMAD, HKBU-MARs V1+, and HiFiMask are abbreviated as D, H, and F, respectively. To facilitate better interpretation, the values for four metrics—HTER, EER, B@A=0.1, and B@A=0.01—are presented as 100 minus their original percentage values. 
	\figurename~\ref{fig6} shows that although existing methods achieve commendable results under the intra-dataset protocols, their performance significantly deteriorates under the challenging cross-dataset protocols. 
	For instance, in the 3DMAD$\rightarrow$HKBU-MARs V1+ protocol (i.e., D to H), the test set contains highly realistic masks not encountered during training, revealing the limited generalization ability of existing methods against 3D mask attacks. 
	In contrast, the proposed method demonstrates balanced and favorable performance across all seven protocols and five metrics. These results show the generalization capability of the proposed method in addressing variations introduced by diverse recording devices, lighting conditions, backgrounds, mask materials, and mask colors. 
	
	The HiFiMask dataset includes a larger number of subjects, making it less susceptible to the effects of random subject selection. Additionally, there are significant data distribution differences between the HiFiMask and 3DMAD datasets, making the HiFiMask$\rightarrow$3DMAD protocol a good representation of real-world scenarios involving 3D mask presentation attack detection. Next, we select this protocol as the representative case and further investigate the working mechanism of the proposed method through a series of component and visualization analysis experiments.
	
	\begin{table}[!tbp]
		%\scriptsize
		\footnotesize
		\rowcolors{1}{}{lightgray}
		\caption{Comparison of different prompts under the HiFiMask$\rightarrow$3DMAD protocol.}
		\label{tabp}
		\centering \begin{tabular}{p{1.5cm}p{0.8cm}<{\centering}p{0.8cm}<{\centering}p{0.8cm}<{\centering}p{1.1cm}<{\centering}p{1.22cm}<{\centering}}
			\hline
			& HTER$\downarrow$&EER$\downarrow$ &AUC$\uparrow$ &B@A=0.1$\downarrow$
			&B@A=0.01$\downarrow$  \\
			%\hline
			$M_1$ & 18.10&5.53&99.75&0.71&1.03
			\\ 
			%\hline
			CoOp~\cite{Zhou_2022}& 12.39&4.55&99.66&0.77&1.09\\
			%\hline
			FLIP~\cite{srivatsan2023flip} & 4.77&1.79&99.83&0.80&1.04  \\
			%\hline		
			Ours&1.04&1.76&99.97&0.05&0.56\\
			\hline		
		\end{tabular}
	\end{table}
	
	\subsection{Comparison with Classical Prompt Learning Methods}
	
	\noindent \textbf{Quantitative comparison.}
	We compare the proposed method with classical prompt learning methods to verify the validity of knowledge-based prompt learning, and the experimental results are shown in Table~\ref{tabp}. ``A photo of'' represents a fixed textual prompt with the vision-language model's parameters fixed. For convenience, we refer to this model as $M_1$. CoOp~\cite{Zhou_2022} uses learnable prompt vectors while keeping the vision-language model's parameters fixed.  FLIP~\cite{srivatsan2023flip} integrates multiple fixed prompts while fine-tuning the vision-language model's parameters. Our method leverages a knowledge-based prompt learning framework, maintaining fixed parameters for the vision-language model.
	
	The EERs of $M_1$ and CoOp are significantly higher compared to the proposed method (5.53\% $vs$ 1.76\% and 4.55\% $vs$ 1.76\%), indicating that even on a development set with a distribution similar to the training set,  these two methods still have significant room for improvement. Furthermore, HTERs of $M_1$ and CoOp are notably higher than their respective EERs (18.10\% $vs$ 5.53\% and 12.39\% $vs$ 4.55\%), revealing that thresholds derived from the development set fail to generalize effectively to the test set. 
	These results demonstrate poor cross-scenario generalization capabilities of these methods, suggesting that simple prompts inadequately capture generalized features from vision-language models for 3D mask presentation attack detection.
	
	While FLIP also utilizes fixed prompts, it differs by fine-tuning the vision-language model's parameters during training, leading to improved detection performance. However, this approach increases computational cost and risks reducing the model's generalization ability, potentially causing overfitting to the training data.
	In contrast, the proposed method integrates prior knowledge into prompt learning while maintaining fixed parameters for the vision-language model, substantially improving HTER (1.04\% $vs$ 12.39\%). The results demonstrate the effectiveness of incorporating prior knowledge into prompts, enabling the transfer of pre-trained vision-language model knowledge for 3D mask presentation attack detection.

	\begin{figure}[!t]
		\centering
		\includegraphics[width=0.5\textwidth]{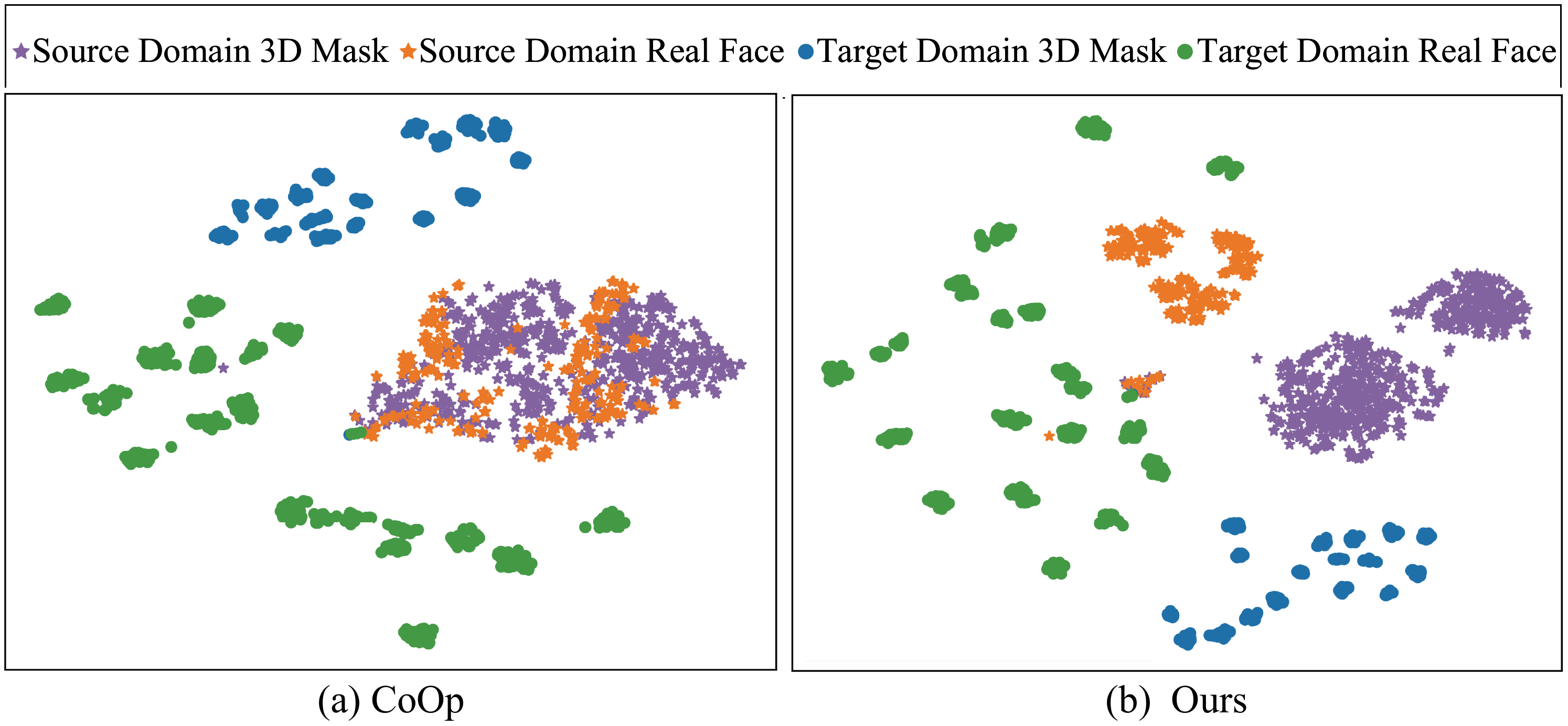}
		\caption{t-SNE plot of the features of (a) CoOp and (b) our method under the HiFiMask$\rightarrow$3DMAD protocol. The source domain represents the training set HiFiMask dataset, and the target domain represents the test set 3DMAD dataset.
		}
		\label{fig11}
	\end{figure}

	\begin{figure*}[!htbp]
		\centering
		\includegraphics[width=1\textwidth,height=0.27\textheight]{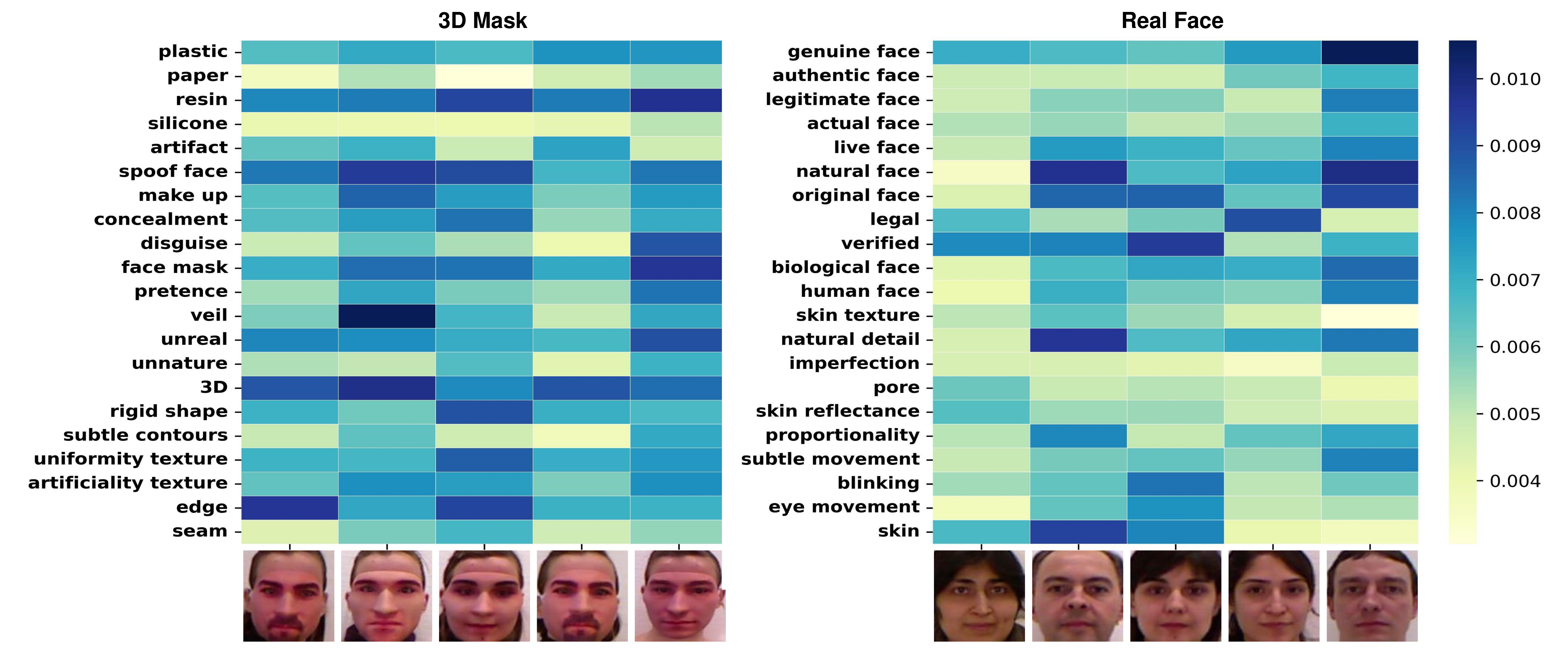}
		\caption{
			 Knowledge graph entity and image correlation analysis under the HiFiMask$\rightarrow$3DMAD protocol. A deeper shade of blue indicates a higher degree of correlation. For both real faces and 3D masks, the knowledge graph prompts attend to entities that are consistent with human intuition and domain knowledge. Compared to class-name prompts, the knowledge graph prompts yield more fine-grained and semantically rich contextual cues.
		}
		\label{fig13}
	\end{figure*}
	
	\noindent \textbf{Comparison of feature visualizations.}
	We use t-SNE to visualize the image feature space of CoOp and the proposed method, with results shown in \figurename~\ref{fig11}. For this visualization, we randomly select 1,000 samples each from the training and test sets. Each point in \figurename~\ref{fig11} represents an individual image.
	The HiFiMask dataset contains highly realistic 3D masks, leading CoOp to cluster 3D masks and real faces from the HiFiMask dataset closely together. This indicates that CoOp’s image encoder, without fine-tuning, struggles to adaptively distinguish between real faces and highly realistic 3D masks. Consequently, the classification boundaries and thresholds established on the training set fail to generalize effectively to the test set.

	In contrast, the proposed method distinctly separates 3D masks and real faces in the training set, achieving tight intra-class clustering and large inter-class distances. This separation enhances the model’s ability to generalize across diverse factors in cross-scenario settings. Furthermore, samples of the same class from both the training and test sets are mapped consistently to the same side of the decision boundary. 
	These results demonstrate that the decision boundary learned by the proposed method on the training set can be successfully applied to the test set, further validating the enhanced generalization ability of the proposed prompt learning approach for detecting 3D mask presentation attacks.

	\subsection{Visualization and Analysis}
        We perform a visual analysis of the correlations between knowledge graph entities and images, as well as category-relevant and category-irrelevant patches within images.
	
        \subsubsection{Analysis of the Correlations between Knowledge Graph Entities and Images}
	  Knowledge graph entities play a pivotal role in knowledge-based prompt learning. 
	Analyzing the effect of different entities in the learning process is beneficial for the dissection of our method.
	We visualize the attention scores between entities and images in the visual-specific knowledge filter $\mathfrak{F}_e$ to achieve this. 
	The results under the HiFiMask$\rightarrow$3DMAD protocol are shown in \figurename~\ref{fig13}. For 3D masks, entities related to material properties, such as ``resin'' and ``plastic'', structure features like ``3D'' and ``edge'' (e.g., between the mask and the face), and associated terms, including ``spoof face'', ``face mask'', and ``evil'' receive significant attention during the decision-making process. 
	In contrast,  for real faces, entities focusing on material characteristics like ``skin'', texture details such as ``natural details'' and ``natural face'', and actions like ``blinking'' and ``subtle movement'', as well as terms like ``genuine face'', ``legal'', and ``verified'' are highly attended. 
	These observations align closely with commonly identified detection cues in 3D mask presentation attack detection. Notably, numerous attended entities represent semantic extensions of class names tailored to the business context of 3D mask detection. 
	Simple class labels such as 3D masks and real faces fail to encapsulate the nuanced information required for robust detection. 
	By explicitly incorporating prior knowledge, our framework generates accurate and comprehensive prompts aligned with practical detection scenarios, significantly improving model performance.
	
	\subsubsection{Analysis of Spurious Correlation Elimination}
	
	\begin{figure}[!htbp]
		\centering
		\includegraphics[width=0.48\textwidth]{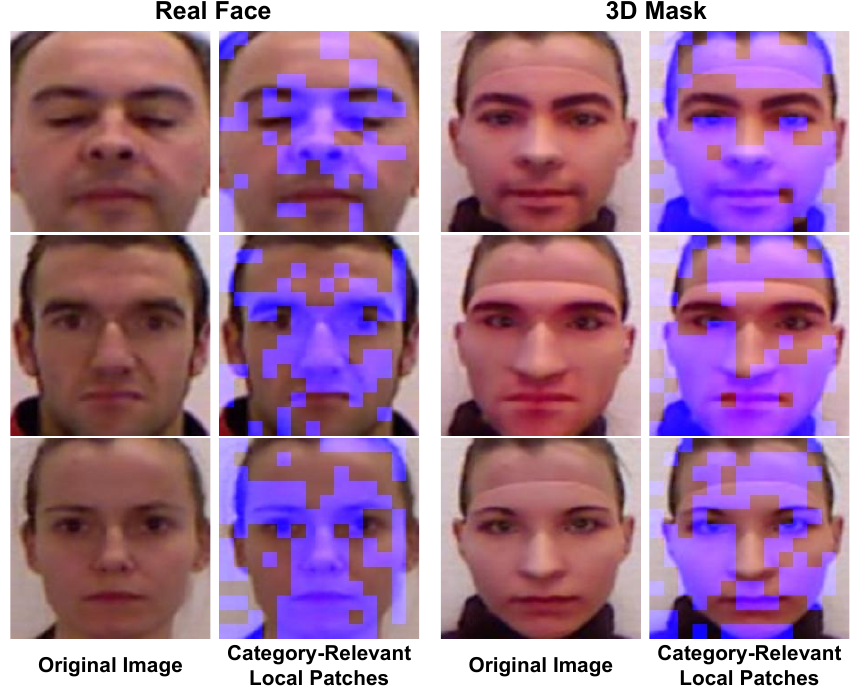}
		\caption{ Visualization of category-relevant local patches that are covered in blue for real faces and 3D masks. The distribution patterns of category-relevant local patches differ significantly between real faces and 3D masks. For a given category, these patches exhibit a relatively consistent distribution across different individuals. 
		}
		\label{fig3}
	\end{figure}
	The spurious correlation elimination learning paradigm enables the detection model to better focus on generic yet category-relevant local patches.
	We perform a visualization analysis of category-relevant and category-irrelevant local patches involved in spurious correlation elimination learning to validate this phenomenon. The results under the HiFiMask$\rightarrow$3DMAD protocol are illustrated in~\figurename~\ref{fig3}. We train the model using the HiFiMask dataset and then visualize the local patches calculated by Equation~\ref{equ4} within the 3DMAD dataset.
	The distribution patterns of category-relevant local patches differ significantly between real faces and 3D masks. For a given category, these patches exhibit a relatively consistent distribution across different individuals. In real faces, the patches are concentrated in areas rich in three-dimensional structures, such as the nose and eyes. In contrast, 3D masks focus on the covered mask areas and the junctions between the mask and the face. 
	These substantial inter-category differences demonstrate the model's strong discriminative capabilities. Furthermore, the consistent discriminative cues within each category underscore the model's good generalization ability.

	\begin{figure*}[!tbp]
	\centering
	\includegraphics[width=0.98\textwidth]{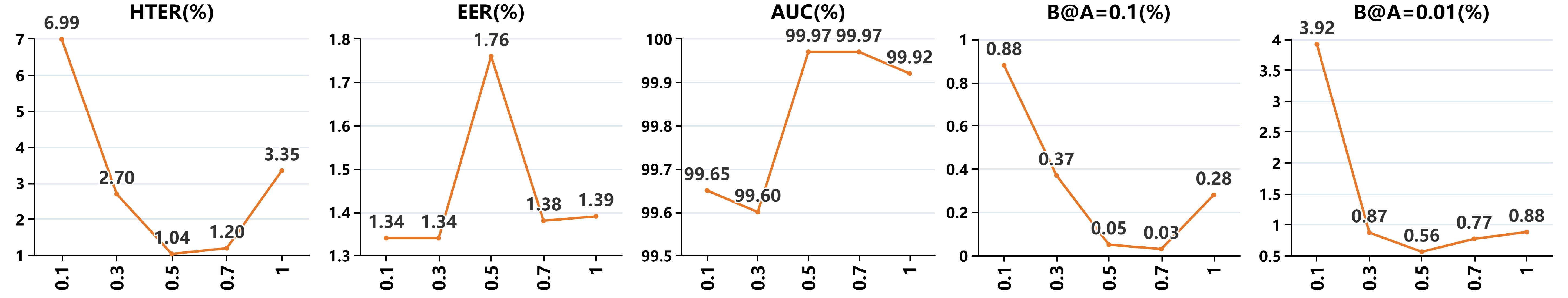}
	\caption{Analysis of different hyperparameter values $\lambda \in \{0.1,0.3,0.5,0.7,1\}$ for spurious correlation elimination learning under the HiFiMask$\rightarrow$3DMAD protocol.}	
	\label{fig8}
	\end{figure*}
	
	\subsection{Ablation Study and Hyperparameter Analysis}
	We evaluate the contribution of each component in the proposed framework to the overall performance. The experimental results under the HiFiMask$\rightarrow$3DMAD protocol are summarized in Table~\ref{tabcom}.
	\subsubsection{Effectiveness of Knowledge Graph Prompt}
	Removing the visual-specific knowledge graph prompt $t_{vskg}^k$ results in HTER performance drops of 88.92\%. 
	The result demonstrates the importance of comprehensive, category-related knowledge expansion in improving cross-scenario generalization performance for 3D mask presentation attack detection.
	When various knowledge components are removed, the EER values exhibit minimal changes, whereas the HTER values show substantial variations.
	Removing the visual-specific entity prompt $\mathfrak{F}_e(t_e^k)$ and visual-specific fine-grained discriminative description $\mathfrak{F}_d(t_d^k)$ results in HTER performance drops of 81.69\% and 75.81\%, respectively. 
	These findings underscore the necessity of both entity prompts and fine-grained descriptions for effective knowledge graph integration. 
	Notably, the pronounced performance decline observed when entity prompts are removed underscores that the structured, high-quality prior knowledge embedded within these prompts plays a crucial role in knowledge graph prompt learning.
	
	Furthermore, the exclusion of the visual-specific knowledge filter $\mathfrak{F}_e$ and $\mathfrak{F}_d$ leads to HTER performance drops of 71.82\% and 83.67\%, respectively. 
	These results demonstrate the effectiveness of weighted integration of knowledge graph entities and descriptions tailored to specific image content. The more pronounced decline caused by removing $\mathfrak{F}_d$ reflects the lower inherent purity of descriptions compared to entities, making targeted refinement based on image content crucial for optimizing detection performance.
	
    \begin{table}[!tbp]
	%\scriptsize
	\footnotesize
	\rowcolors{1}{}{lightgray}
	%\captionsetup{justification=justified, singlelinecheck=false}
	\caption{Component analysis results under the  HiFiMask$\rightarrow$3DMAD protocol. The performance variations resulting from the individual removal of visual-specific knowledge graph prompt $t_{vskg}^k$, visual-specific entity prompt $\mathfrak{F}_e(t_e^k)$, visual-specific fine-grained discriminative description $\mathfrak{F}_d(t_d^k)$, visual-specific entity filter $\mathfrak{F}_e$, visual-specific description filter $\mathfrak{F}_d$, and spurious correlation elimination regularization $\ell_{sce}$ are compared.}
	\label{tabcom}
	\centering
	\begin{tabular}{p{1.5cm}p{0.9cm}<{\centering}p{0.7cm}<{\centering}p{0.8cm}<{\centering}p{1.22cm}<{\centering}p{1.22cm}<{\centering}}
		\hline
		%& \multicolumn{5}{c|}{HiFiMask$\rightarrow$3DMAD}   \\
		%\hline
		& HTER$\downarrow$&EER$\downarrow$ &AUC$\uparrow$ &B@A=0.1$\downarrow$
		&B@A=0.01$\downarrow$  \\
		%\hline
		w/o $t_{vskg}^k$&9.39&1.32&99.75&0.82&1.23\\
		%\hline
		w/o $\mathfrak{F}_e(t_e^k)$ &5.68&1.48&99.88&0.49&0.86\\
		%\hline
		w/o $\mathfrak{F}_d(t_d^k)$ & 4.30&1.34&99.90&0.38&0.87 \\
		%\hline
		w/o $\mathfrak{F}_e$ &3.69&1.49&99.71&0.77&1.00\\
		%\hline
		w/o $\mathfrak{F}_d$ &6.37&1.30&99.78&0.82&1.32\\
		%\hline
		w/o $\ell_{sce}$ &2.41&1.76&99.90&0.37&0.85\\
		%\hline
		Ours&1.04&1.76&99.97&0.05&0.56\\
		\hline	
	\end{tabular}
    \end{table}	
	
	\subsubsection{Effectiveness of Spurious Correlation Elimination Regularization.}
	Removing the spurious correlation elimination regularization $\ell_{sce}$ results in HTER performance drops of 56.85\%.
	The result reveals that eliminating spurious correlations facilitates the causal alignment of knowledge-based prompts with category-relevant local patches, which is crucial for the improvement of detection performance. 
	The minimal change in EER performance, along with the decline in HTER performance, also suggests that the category-relevant local patches exhibit strong cross-scenario generalization capabilities.

	\subsubsection{Analysis of the Hyperparameter for Spurious Correlation Elimination Regularization }
	
	Spurious correlation elimination learning involves a hyperparameter $\lambda$ that balances the weight between spoof-relevant discriminant regularization ($\ell_{srd}$) and spurious correlation elimination regularization ($\ell_{sce}$). We evaluate the impact of different $\lambda$ values on detection performance, with the experimental results presented in \figurename~\ref{fig8}.
	As $\lambda$ increases from 0.1 to 1, HTER initially decreases and then increases, with the optimal performance across all five metrics observed at $\lambda = 0.5$. These results indicate that the influence of the spurious correlation elimination regularization $\ell_{sce}$ should be moderate during prompt learning. Based on these findings, $\lambda$ is set to 0.5 in this paper.

	\begin{table}[!tbp]
		%\scriptsize
		\footnotesize
		\rowcolors{1}{}{lightgray}
		\caption{ Comparison of different visual-language models as the backbone under the HiFiMask$\rightarrow$3DMAD protocol.}
		\label{tabback}
		\centering \begin{tabular}{p{2.8cm}p{0.7cm}<{\centering}p{0.4cm}<{\centering}p{0.5cm}<{\centering}p{1cm}<{\centering}p{1.1cm}<{\centering}}
			\hline
			%& \multicolumn{5}{c|}{HiFiMask$\rightarrow$3DMAD}   \\
			%\hline
			& HTER$\downarrow$&EER$\downarrow$ &AUC$\uparrow$ &B@A=0.1$\downarrow$
			&B@A=0.01$\downarrow$  \\
			%\hline
			CLIP ViT-B/16 &1.04&1.76&99.97&0.05&0.56\\
			CLIP ViT-L/14&0.77&0.88&99.97&0.03&0.75\\
			CLIP ViT-L/14@336px&0.90&1.15&99.99&0.00&0.20\\
			SigLIP ViT-B/16&0.86&1.06&99.99&0.01&0.08\\
			\hline	
		\end{tabular}
	\end{table}

	 \subsubsection{Analysis of the Backbone Visual-Language Model }

	We evaluate the effectiveness of our method when integrated with different backbone vision-language models, including CLIP ViT-B/16, CLIP ViT-L/14, CLIP ViT-L/14@336px, and SigLIP ViT-B/16. The corresponding results are presented in Table~\ref{tabback}. 
	Our approach transfers the general capabilities of vision-language models to enhance the performance of 3D mask presentation attack detection.
	From the results obtained using CLIP ViT-B/16, CLIP ViT-L/14, and SigLIP ViT-B/16, we observe a consistent trend: more powerful vision-language models lead to improved detection performance. Notably, CLIP ViT-L/14 and CLIP ViT-L/14@336px share the same model architecture but differ in input image resolution. A comparison between these two variants reveals that increasing the input resolution does not yield performance gains in our setting.
	This phenomenon may be attributed to the characteristics of 3D mask presentation attack detection datasets, in which face images are typically pre-aligned to a resolution of $128 \times 128$. Simply upsampling to $336 \times 336$ does not introduce additional spoof patterns or enhance fine-grained details, thus offering limited benefit to the model.
	
   \begin{figure}[!tbp]
		\centering
		\includegraphics[width=0.5\textwidth]{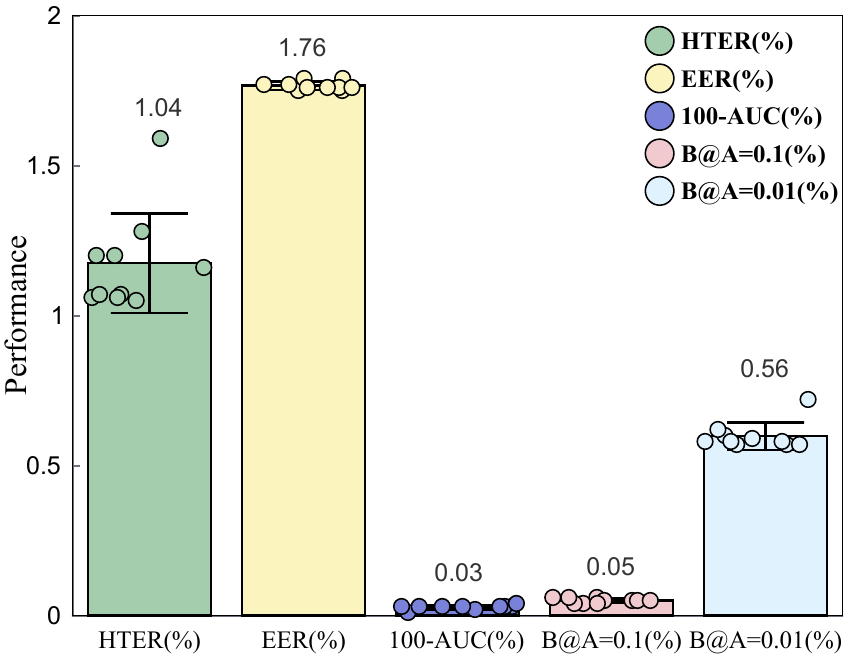}
		\caption{
			Performance across five evaluation metrics under the HiFiMask $\rightarrow$ 3DMAD protocol with random removal of a single entity from the knowledge graph (repeated over ten trials).
		Each bar represents the average performance across ten independent trials for a given metric, while the error bars denote the range of observed performance variations. For reference, the performance without entity removal is annotated on each bar. Experimental results demonstrate that the performance variations across all five evaluation metrics remain consistently small, indicating the strong robustness of our method to the removal of a single entity from the knowledge graph.
		}
		\label{fig9}
	\end{figure}
     
	\subsubsection{Robustness Analysis of the Number of Entities } 
	 Entities play a pivotal role in knowledge-based prompt learning. To further investigate their impact, we analyze how the quantity of entities influences the performance of 3D mask presentation attack detection. \figurename~\ref{fig9} presents the model performance under the HiFiMask $\rightarrow$ 3DMAD protocol after randomly removing one entity from the knowledge graph, repeated over ten independent trials. Notably, when an entity is removed, all descriptions associated with its corresponding triples are also excluded.
	Although removing different entities results in slight performance degradation, the overall fluctuation across the five evaluation metrics remains minimal, indicating that the proposed method is robust to the removal of a single entity. This robustness can be attributed to the model's ability to leverage fine-grained discriminative descriptions, which draw on prior knowledge extracted from large-scale language models. Such capability enables the model to effectively expand the removed entities. As a result, the information loss incurred by removing a single entity can often be compensated by the contextual expansion of related entities, thereby leading to negligible performance variation.
	
\begin{figure}[!tbp]
		\centering
		\includegraphics[width=0.5\textwidth]{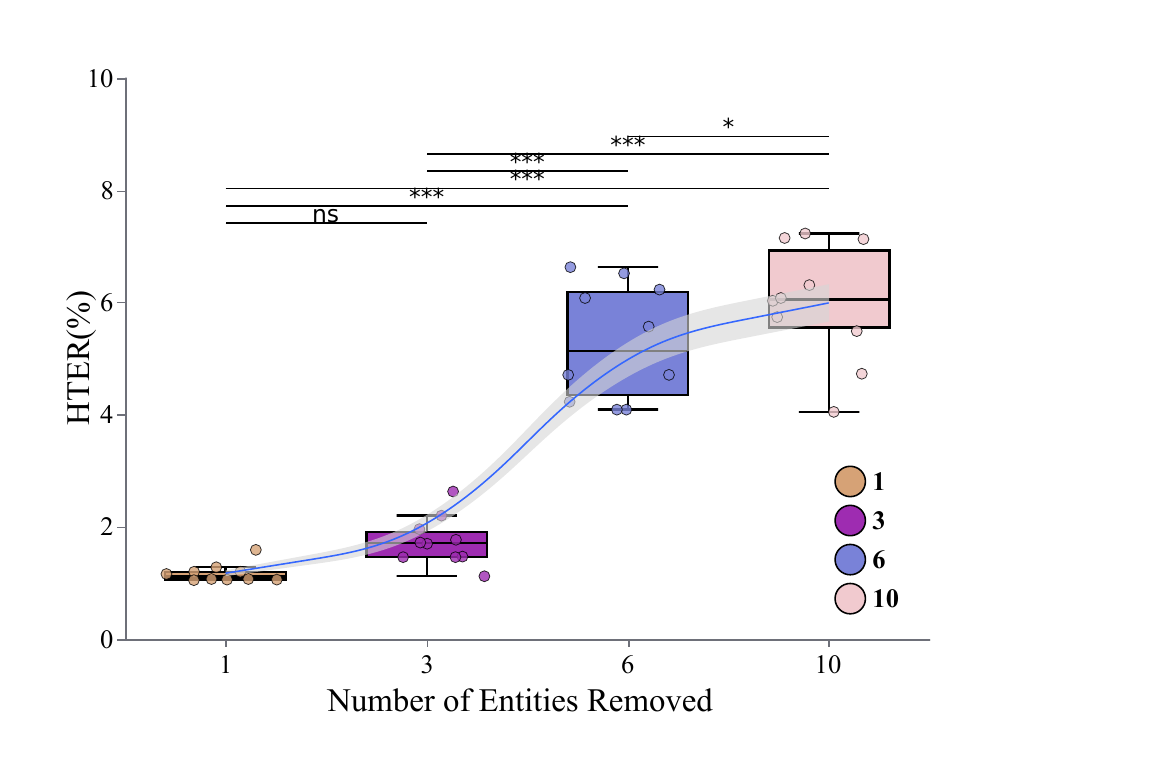}
		\caption{
		HTER performance under the HiFiMask $\rightarrow$ 3DMAD protocol with varying numbers of randomly removed knowledge graph entities (ten repeated trials for each setting).
		Box plots show HTER distributions for each removal level, with statistical significance between different removal levels indicated above ("ns" for non-significance, asterisks for increasing significance). The ribbon error line illustrates the HTER mean trend and variability across removal levels. Overall, model performance declines as more entities are removed, especially when the remaining entities can no longer compensate. This underscores the critical role of knowledge graph entities in model effectiveness and stability.
		}
		\label{fig10}
    \end{figure}

	To further evaluate the model sensitivity to entity quantity, we perform pairwise differential t-tests on the performance distributions obtained under four conditions: randomly removing one, three, six, and ten entities, respectively, under the HiFiMask $\rightarrow$ 3DMAD protocol. \figurename~\ref{fig10} summarizes the results.
	The box plots illustrate the distribution of HTER scores across repeated trials for each level of entity removal. As the number of removed entities increases, we observe a gradual decline in median performance, alongside an expansion in both the interquartile range and whisker lengths, indicating increased variability. Pairwise statistical significance results are annotated above the plots, where "ns" denotes no statistically significant difference, and an increasing number of asterisks (*) indicates stronger significance. It is evident that the performance difference between removing one and three entities is marginal, whereas the performance degradation becomes statistically significant when six or more entities are removed.
	In addition, the ribbon error line further visualizes the HTER variation trend across removal levels. The central line, representing the mean HTER, increases steadily as more entities are removed, while the ribbon width grows accordingly, reflecting growing performance variability.
	Overall, the performance of the model deteriorates with the incremental removal of entities, particularly when the number of removals increases beyond the point where compensation through remaining entities is no longer effective. These findings underscore the critical role of knowledge graph entities in enhancing the performance of 3D mask presentation attack detection.

	 \subsection{Generalization to Diverse Attack Types.}

	\begin{table}[!tbp]
		%\scriptsize
		\footnotesize
		\rowcolors{1}{}{lightgray}
		\caption{ Cross-dataset diverse physical attack types evaluation results under the CS$\rightarrow$W and C$\rightarrow$H protocol.}
		\label{tabopen1}
		\centering 
		\begin{tabular}{p{1.5cm}p{0.8cm}<{\centering}p{0.8cm}<{\centering}|p{1.7cm}p{0.8cm}<{\centering}p{0.8cm}<{\centering}
			}
			\hline
			Method& \multicolumn{2}{c|}{CS$\rightarrow$W}   &Method& \multicolumn{2}{c}{C$\rightarrow$H}   \\
			& HTER$\downarrow$& AUC$\uparrow$  && HTER$\downarrow$& AUC$\uparrow$  \\
			ViT~\cite{huang2022adaptive} &21.04 &89.12  &ConvMLP~\cite{wang2022conv} &48.02 &53.45\\
			CLIP-V~\cite{radford2021learning} &20.00 &87.72  &PipeNet~\cite{yang2020pipenet} &49.94 &51.60\\
			CLIP~\cite{radford2021learning}&17.05&89.37   &FeatherNet~\cite{zhang2019feathernets} &57.70 &38.54 \\
			FGPL~\cite{hu2024fine}  &14.05  &92.65 &FlexModal~\cite{yu2023flexible} &57.70 &66.17 \\
			CoOp~\cite{Zhou_2022} &9.52& 90.49  &FaceBagNet~\cite{shen2019facebagnet} &47.79 &53.47 \\ 
			CFPL~\cite{liu2024cfpl}  &9.04 &96.48   &ViT-S/16~\cite{dosovitskiy2020image} &49.00 &52.33 \\
			S-CPTL~\cite{guo2024style}& 8.99 &94.01  &mmFAS~\cite{chen2025mmfas} &36.92 &69.85 \\
			Ours&\textbf{7.76}&\textbf{98.03} &Ours&\textbf{15.43}&\textbf{92.68}\\
			\hline
		\end{tabular}
	\end{table}
    
	In addition to evaluating performance on datasets containing only 3D mask attacks, we further design open-set scenarios involving diverse physical attack types and digital adversarial attacks to assess the generalization capability of our method.
	
	\textbf{Generalization to diverse physical attack types.} The evaluation experiments on diverse physical attack types are conducted based on the CASIA-SURF (S)~\cite{zhang2020casia}, CASIA-SURF CeFA (C in short)~\cite{li2020casia}, WMCA (W in short)~\cite{george2019biometric}, and HQ-WMCA (H in short)~\cite{heusch2020deep} datasets. The WMCA and HQ-WMCA datasets include 7 and 10 different attack types, respectively, such as print attacks, replay attacks, mannequin heads, paper mask, rigid mask, flexible mask, glasses, wigs, tattoo, and makeup. We construct the CS $\rightarrow$ W protocol as S-CPTL~\cite{guo2024style}, using the attack-rich WMCA dataset as the test set and the CASIA-SURF and CASIA-SURF CeFA datasets as the training data. Furthermore, we introduce a more challenging C $\rightarrow$ H protocol as mmFAS~\cite{chen2025mmfas}, which reduces the number of datasets in the training data while increasing the diversity of attack types in the test data, to further verify the generalization ability of our method in more universal scenarios.
	
	The experimental results are presented in Table~\ref{tabopen1}. As shown, our method outperforms existing SOTA methods on the HTER metric by 13.68\% and 58.20\% under the two protocols. This substantial improvement demonstrates the strong generalization capability of our method to diverse attack types. On the left of Table~\ref{tabopen1}, the comparison methods are mostly prompt learning-based approaches, all using the same backbone vision-language model, CLIP ViT-B/16, as ours. The significant gains under the CS $\rightarrow$ W protocol further highlight the effectiveness of our knowledge graph prompts. On the right of Table~\ref{tabopen1}, the evaluated methods are primarily multi-modal fusion approaches. Notably, our method leverages only visible-light modality images for both training and testing, yet still achieves considerable improvements under the C $\rightarrow$ H protocol. This demonstrates that our approach can effectively generalize to cross-dataset, diverse attack types using only a single modality and a single training dataset.

	\begin{table}[!tbp]
		%\scriptsize
		\footnotesize
		\rowcolors{1}{}{lightgray}
		\caption{ Cross-dataset adversarial attack evaluation results under the HiFiMask$\rightarrow$3DMAD protocol.}
		\label{tabadv}
		\centering \begin{tabular}{p{1.4cm}p{1cm}<{\centering}p{0.7cm}<{\centering}p{1cm}<{\centering}p{1cm}<{\centering}p{1.2cm}<{\centering}}
			\hline
			Method& HTER$\downarrow$&EER$\downarrow$ &AUC$\uparrow$ &B@A=0.1$\downarrow$
			&B@A=0.01$\downarrow$  \\
			FASTEN~\cite{cao2024flow}&44.71&47.06&52.25&97.06&100.00\\
			FLIP~\cite{srivatsan2023flip}&4.12&\textbf{2.21}&99.99&0.00&\textbf{0.00}\\
			Ours&\textbf{2.65}&7.15&\textbf{99.99}&\textbf{0.00}&1.76\\
			\hline	
		\end{tabular}
	\end{table}

	\textbf{Generalization to digital adversarial attacks.} The evaluation experiments on digital adversarial attacks are conducted based on the HiFiMask and 3DMAD datasets.
	We use the HiFiMask dataset as the training set. For the test set, we construct a combination of real face images from the 3DMAD dataset and adversarial attack samples generated by AT3D~\cite{yang2023towards} using the real face images from the 3DMAD dataset as original images.
	The experimental results are presented in Table~\ref{tabadv}. The performance of the FASTEN method is obtained using the officially released pretrained model. %~\footnote{https://github.com/yuxincao22/FASTEN}. 
	As shown, FASTEN exhibits poor generalization to adversarial samples. In contrast, prompt learning-based methods demonstrate significantly better generalization ability to adversarial attacks. This indicates that leveraging prompt learning to tap into the general knowledge of vision-language models is beneficial for handling adversarial attacks.
	Compared to the simple category name prompts of FLIP, our method achieves a 37.16\% improvement on the HTER metric, further validating the effectiveness of the proposed knowledge graph prompts.

    \begin{table}[!tbp]
		%\scriptsize
		\footnotesize
		\rowcolors{1}{}{lightgray}
		\caption{ Comparative results of computational complexity on a single GeForce RTX 3090 GPU.}
		\label{tabcompu}
		\centering \begin{tabular}{p{1.4cm}p{1.9cm}<{\centering}p{1.9cm}<{\centering}p{1.9cm}<{\centering}}
			\hline
			Method& Inference Times&Training Parameters &Frozen Parameters \\
			FASTEN~\cite{cao2024flow}&23.06 ms&39.8M&162.5M\\
			Ours&\textbf{3.91 ms}&\textbf{12.7M}&\textbf{149.6M}\\
			\hline	
		\end{tabular}
    \end{table}

	\subsection{Computational Complexity Analysis.}
	
	We compare the computational complexity of our method with the baseline method FASTEN in terms of inference time and model parameters. The experimental results are presented in Table~\ref{tabcompu}.
	Our method is single-frame based, with an average inference time of 3.91 ms, while FASTEN is a multi-frame based method, with an average inference time of 23.06 ms. 
	We freeze the CLIP backbone (149.6M parameters) during training and optimize only lightweight prompt parameters and filters, totaling 12.7M parameters. In contrast, FASTEN involves multi-frame processing and spatiotemporal modeling, requiring training a FlowNetS (38M parameters) and a MobileNetV3-Small (1.8M parameters) backbone, along with a pre-trained FlowNet2.0 (162.5M parameters) used for generating ground-truth optical flow labels prior to training.
	All these results indicate that our method holds an advantage in terms of computational complexity.

	\section{Conclusion}
	
	This paper introduces a knowledge-based prompt learning framework for 3D mask presentation attack detection. By integrating expert knowledge through knowledge graphs and causal graphs, the framework effectively adapts pre-trained vision-language models for 3D mask presentation attack detection. Extensive experiments demonstrate that the proposed method achieves state-of-the-art performance in both intra- and cross-scenario evaluations. 
	However, the proposed method has certain limitations. The knowledge graph presented in this paper is manually constructed, which inherently restricts the number of entities and relationships it can encompass. Additionally, variations in expert experiences may lead to potential discrepancies. In future work, we aim to develop an automated approach for constructing a more unified and comprehensive knowledge graph.

	% or
	
	
	
	\ifCLASSOPTIONcompsoc
	%	\section*{Acknowledgments}
	\else
	%	\section*{Acknowledgment}
	\fi
    
	%\section*{Acknowledgment}
	%This work was supported in part by the National Key Research and Development Program of China (Grant No. 2022YFC3310400), in part by the Natural Science Foundation of China (Grant Nos. U23B2054, 62076240, 62102419, 62276263 and 62406133), in part by the Beijing Municipal Natural Science Foundation (Grant No. 4222054), in part by the Natural Science Foundation of Hunan Province (Grant No.2024JJ6389 and 2024JJ7437), in part by the Hengyang Science and Technology Plan Project (Grant No.202330046190), and in part by Interdisciplinary Research Program in Medicine and Engineering, The First Affiliated Hospital of University of South China (Grant No.IRP-M$\&$E-2025-12).
	
	\ifCLASSOPTIONcaptionsoff
	\newpage
	\fi
	
	\bibliographystyle{IEEEtran}
	\bibliography{ref2}
	
	\vspace{-15mm}
	\begin{IEEEbiography}
		[{\includegraphics[width=1in,height=1.25in,clip,keepaspectratio]{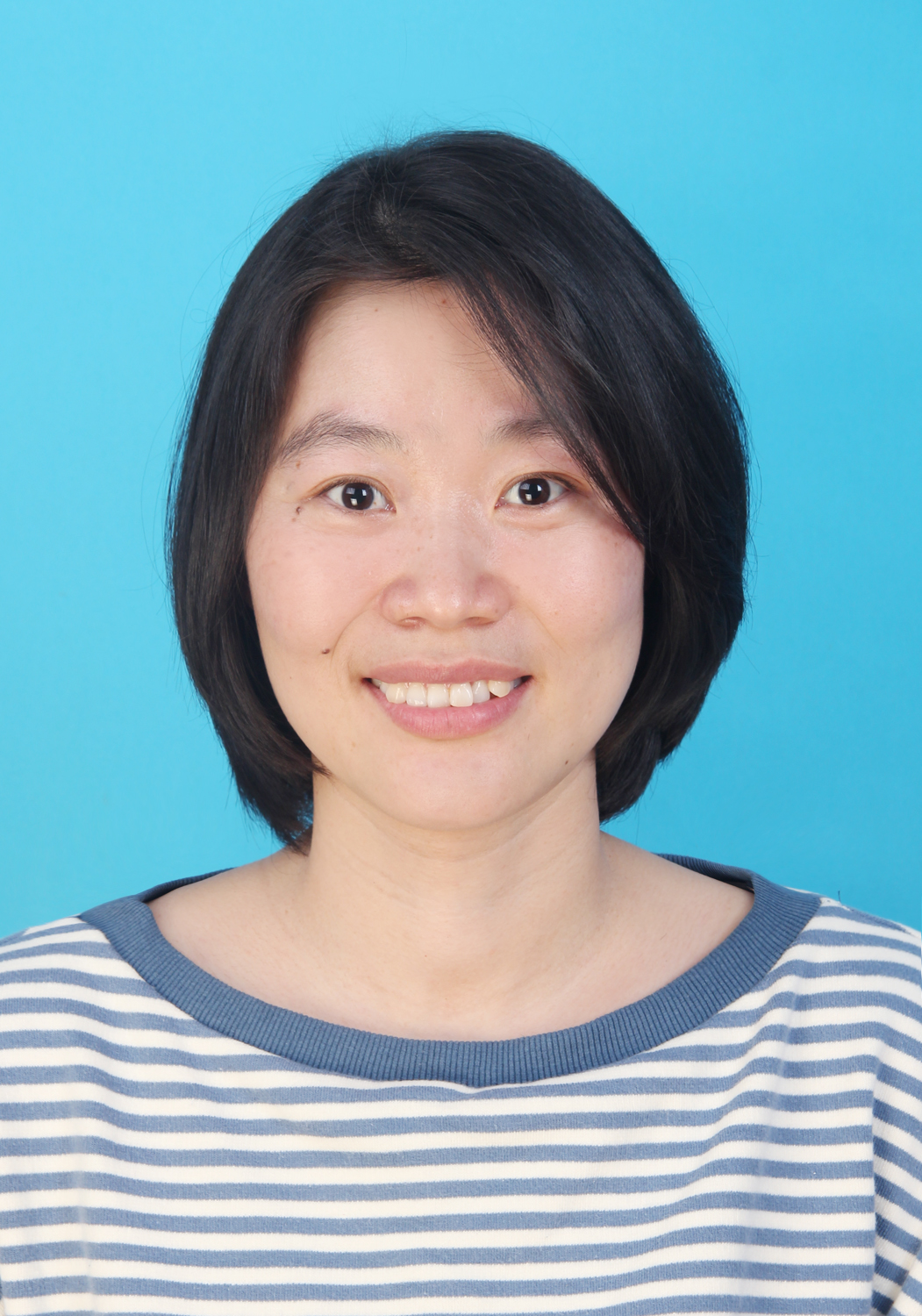}}]
		{Fangling Jiang} received the B.E. and M.S. degrees from Tianjin University in 2009 and 2012, respectively, and the Ph.D. degree from University of Chinese Academy of Sciences in 2021. She is an Assistant Professor with the School of Computer Science, University of South China. Her research interests include face anti-spoofing, transfer learning, and computer vision.
	\end{IEEEbiography}
	
	\vspace{-15mm}
	\begin{IEEEbiography}
		[{\includegraphics[width=1in,height=1.25in,clip,keepaspectratio]{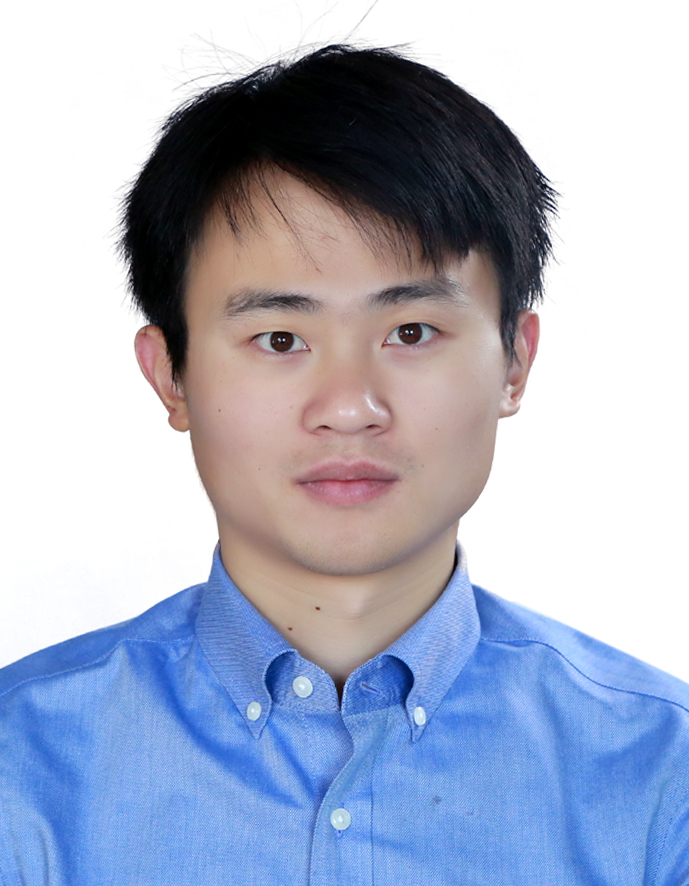}}]
		{Qi Li} received the B.E. degree from China University of Petroleum in 2011, the Ph.D. degree from the Institute of Automation, Chinese Academy of Sciences (CASIA) in 2016. 
		He is an Associate Professor with the New Laboratory of Pattern Recognition (NLPR), State Key Laboratory of Multimodal Artificial Intelligence Systems (MAIS), CASIA. 
		His research interests include face recognition, computer vision, and machine learning.
	\end{IEEEbiography}

	\vspace{-15mm}
	\begin{IEEEbiography}
		[{\includegraphics[width=1in,height=1.25in,clip,keepaspectratio]{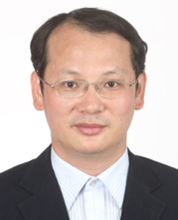}}]
		{Bing Liu} is a Professor of Computer Science and Technology at University of South China. Liu serves as a project head of Industry Academia Collaborative Education of the Ministry of Education in 2024. He is the executive director of the Hunan Provincial University Network Association. His research interests include computer vision and machine learning.
	\end{IEEEbiography}
	
	\vspace{-15mm}
	\begin{IEEEbiography}[{\includegraphics[width=1in,height=1.25in,clip,keepaspectratio]{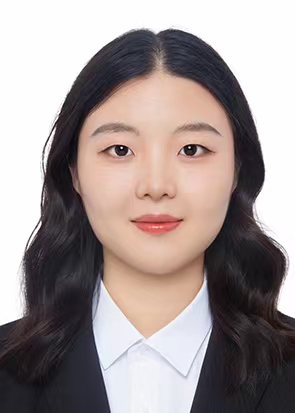}}]{Weining Wang}
		received her B.E. degree from North China Electric Power University in 2015 and the Ph.D. degree from University of Chinese Academy of Sciences (UCAS) in 2020. She is currently an Associate Professor at Institute of Automation, Chinese Academy of Sciences (CASIA). Her research interests include pattern recognition, computer vision and multimodal understanding and generation.
	\end{IEEEbiography}

	\vspace{-15mm}
	\begin{IEEEbiography}[{\includegraphics[width=1in,height=1.25in,clip,keepaspectratio]{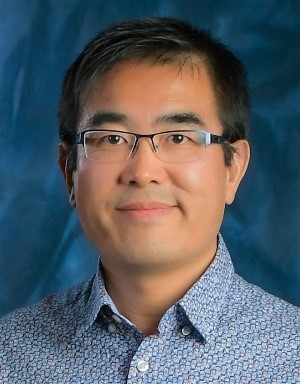}}]{Caifeng Shan}
		received the B.Eng. degree from the University of Science and Technology of China, the M.Eng. degree from the Institute of Automation, Chinese Academy of Sciences, and the Ph.D. degree from Queen Mary, University of London. His research interests include computer vision, pattern recognition, medical image analysis, and related applications. He has authored 150 papers and 80 patent applications. He has served as Associate Editor for journals including IEEE Journal of Biomedical and Health Informatics and IEEE Transactions on Circuits and Systems for Video Technology. He is a Senior Member of IEEE.
	\end{IEEEbiography}
	
	\vspace{-12mm}
	\begin{IEEEbiography}[{\includegraphics[width=1in,height=1.25in,clip,keepaspectratio]{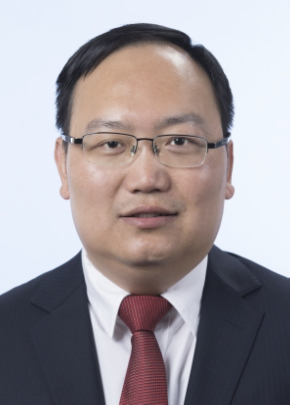}}]{Zhenan Sun}
		received the Ph.D. degree from the Institute of Automation, Chinese Academy of Sciences (CASIA) in 2006.
		He is a professor at New Laboratory of Pattern Recognition (NLPR), State Key Laboratory of Multimodal Artificial Intelligence Systems (MAIS), CASIA.
		His current research interests include biometrics, pattern recognition, and computer vision. He is a fellow of the IAPR, and an Associate Editor of the IEEE Transactions on Biometrics, Behavior, and Identity Science.
	\end{IEEEbiography}
	
	\vspace{-12mm}
	\begin{IEEEbiography}
		[{\includegraphics[width=1in,height=1.25in,clip,keepaspectratio]{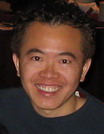}}]
		{Ming-Hsuan Yang} is a Professor of Electrical
		Engineering and Computer Science at University
		of California, Merced. Yang serves as a program
		co-chair of IEEE International Conference on
		Computer Vision (ICCV) in 2019.
		He received the Best Paper Award at ICML 2024, Longuet-Higgins Prize at CVPR 2023, NSF CAREER award in 2012
		and Google Faculty Award in 2009. He is a Fellow of the IEEE, ACM, and AAAI.
	\end{IEEEbiography}

\end{document}